\documentclass[letterpaper]{article} 
\usepackage{aaai2027}  
\usepackage[hyphens]{url}  
\usepackage{graphicx} 
\def\UrlFont{\rm}  
\usepackage{natbib}  
\usepackage{caption} 
\usepackage{algorithm}
\usepackage{algorithmic}
\usepackage{amsmath}
\usepackage{multirow}
\usepackage{amssymb}
\usepackage{pifont}
\newcommand{\cmark}{\ding{51}}
\newcommand{\xmark}{\ding{55}}

\usepackage{newfloat}
\usepackage{listings}
\DeclareCaptionStyle{ruled}{labelfont=normalfont,labelsep=colon,strut=off} 
\floatstyle{ruled}
\newfloat{listing}{tb}{lst}{}
\floatname{listing}{Listing}

\usepackage{booktabs}

\title{When Generative Replay Meets Evolving Deepfakes: Dual Confusion-Aware Regularization for Incremental Face Forgery Detection}
\author{
Hao Shen$^{1*}$, Jikang Cheng$^{2*}$, Renye Yan$^{2*}$, Zhongyuan Wang$^{3}$, Wei Peng$^{4}$, Baojin Huang$^{1\dagger}$ \\[0.2cm]
$^{1}$Huazhong Agricultural University, $^{2}$Peking University \\
$^{3}$Wuhan University, $^{4}$Stanford University \\[0.2cm]
\small $^{*}$These authors contributed equally to this work \\
\small $^{\dagger}$Correspondence to: huangbaojin@mail.hzau.edu.cn
}
\nocopyright
\affiliations{

}

\begin{document}

\maketitle

\begin{abstract}
  The rapid advancement of face generation techniques has introduced an increasing variety of forgery methods, making incremental deepfake detection essential for maintaining robust detectors. While generative replay provides a promising solution to alleviate catastrophic forgetting without storing historical data, its effectiveness is hindered by \textbf{domain confusion} between generated samples and real data. We observe that replay samples fall into two categories: when the replay generator closely resembles the newly introduced forgery model, generated real samples overlap with the fake domain and become \textbf{domain-risky}; when the generator differs significantly, generated samples maintain clearer domain separation and can be treated as \textbf{domain-safe}. To address this challenge, we propose a Dual \textbf{C}onfusion-\textbf{A}ware \textbf{RE}gularization strategy, termed \textbf{Dual-CARE}. A Domain-aware Confusion Score (DC Score) is introduced to quantify domain confusion and \textbf{dual-modulate} the optimization of both replay generators and the incremental detector. Guided by DC Score, replay generators are updated to better approximate previous-task distributions, while the detector adopts different supervision strategies: domain-safe samples are directly supervised, whereas domain-risky samples are regulated using a Relative Separation Loss to balance supervision and confusion. Extensive experiments demonstrate that Dual-CARE effectively exploits generative replay and improves incremental deepfake detection under evolving forgery scenarios.
\end{abstract}

\section{Introduction}
\label{sec:intro}
The misuse of deepfakes severely threatens online trust. While current detection methods \cite{sun2022dual,cao2022end,huang2023implicit,yan2024transcending,cheng2024can,cheng2025ed} focus on leveraging existing samples to improve generalization, they struggle against rapidly evolving forgery techniques. Moreover, retraining models on combined datasets introduces high computational costs and privacy risks. Therefore, an incremental learning strategy is essential to enable continuous model adaptation.

\begin{figure}[t]
    \centering
    \includegraphics[width=1\linewidth]{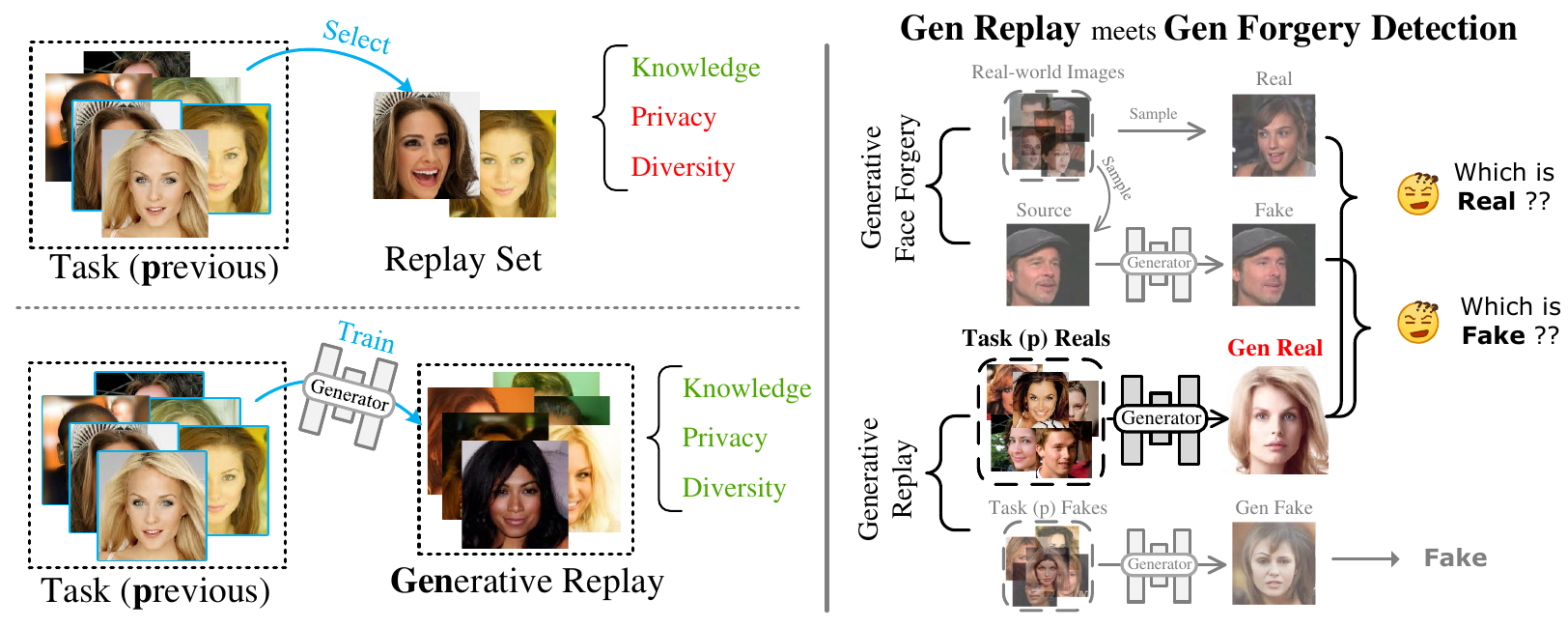}
    \caption{Left: Comparison between traditional sample replay and generative replay. Right: The challenge of applying generative replay to forgery detection, as both Gen-Real and Fake are generated through similar processes, making it difficult for the detector to distinguish real from fake.}
    \label{fig:intro}
\end{figure}

To preserve historical task information, current incremental learning-based forgery detection methods typically utilize various sample replay strategies, such as central/hard sample replay \cite{pan2023dfil}, adversarial perturbations \cite{sun2025continual}, mixed prototypes \cite{tian2024dynamic}, and sparse uniform replay \cite{cheng2025stacking}. However, as illustrated in Fig. \ref{fig:intro}, replaying original samples faces two major bottlenecks: insufficient data diversity due to restricted storage, and potential privacy and security risks. We provide a detailed discussion on \textit{\textbf{Related Work}} in Appendix. Although generative replay avoids these issues by synthesizing samples instead of storing originals, it introduces a fundamental challenge: since the goal of forgery detection is to distinguish real data from synthetic, can these generated "real" samples truly serve as authentic data for model training? Motivated by this question, we make the first attempt to explore the feasibility of generative replay in forgery detection.

\begin{figure}
    \centering
    \includegraphics[width=1\linewidth]{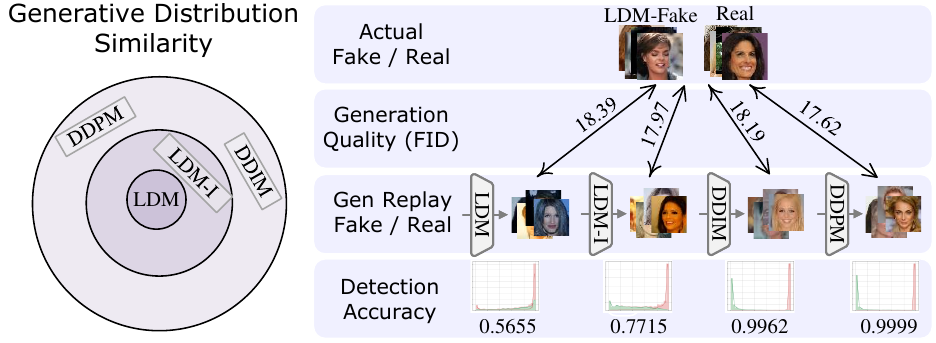}
    \caption{Influence of distribution similarity on generative replay. LDM-I (ODE version~\cite{song2020denoising} of LDM) generates images more similar to LDM than image-level DDPM (using SDE) and DDIM. All replay generators are controlled to have comparable FID scores, ensuring similar generation quality. However, under such a constrained condition, when the replay distribution is closer to the original fake distribution, detection accuracy drops, indicating that generative replay performs better when the replay generator differs more from the deepfake generator.}
    \label{fig:dist-intro}
\end{figure}

As shown in Fig. \ref{fig:dist-intro}, we study generative replay using LDM-generated fake faces together with real faces to train a detection model. Four generators, LDM \cite{rombach2022high}, LDM-I \cite{rombach2022high}, DDIM \cite{song2020denoising}, and DDPM \cite{ho2020denoising}, are used to replay both fake and real distributions, with comparable FID values maintained to ensure fairness. The results show that when the replay generator closely resembles the fake model, such as LDM or LDM-I, the generated samples are poorly distinguished as real or fake. In contrast, generators that differ more from the fake model, such as DDIM or DDPM, achieve accurate detection. These observations suggest that generative replay can be effective in some scenarios but unreliable in others. We refer to the reliable cases as \textbf{domain-safe} and the unreliable ones as \textbf{domain-risky}. Understanding the underlying factors that determine whether a scenario is domain-safe or domain-risky is crucial for designing robust replay strategies.

To fully exploit generative replay for Generated Face Forgery Detection, we propose a Dual \underline{C}onfusion-\underline{A}ware \underline{RE}gularization (Dual-CARE) strategy. The key challenge lies in effectively utilizing replayed samples while mitigating the domain confusion introduced by generative artifacts. To address this, we first introduce a Domain-aware Confusion Score (DC Score) to quantify the degree of domain confusion during training, which is then used to \textbf{dual-modulate} the loss tradeoff in both generator updating and detector learning. Based on this score, we train a pair of updating replay generators that progressively simulate the distributions of all previous-task domains. A domain-aware fast training scheme guided by the DC Score is adopted to enhance the stability and efficiency of generator updating. For detector training, replayed samples are further categorized into “domain-safe” and “domain-risky” groups. The domain-safe samples are directly used for supervised training, while for domain-risky samples we introduce a Relative Separation Loss (RS Loss) to balance the supervision between informative forgery cues and potential confusion introduced by generative artifacts. Extensive experiments demonstrate that Dual-CARE effectively improves detection performance across different replay generators while alleviating the negative effects of domain overlap in incremental deepfake detection. Our contributions can be summarized as follows:
\begin{itemize}
    \item We are the first to explore the feasibility of applying generative replay to generated face forgery detection, analyzing how the overlap between generated “real” and fake samples introduces domain risks that challenge traditional replay assumptions.
    \item We propose Dual-CARE to utilize replayed samples while mitigating interference from generative artifacts, introducing RS Loss and DC Score to adaptively balance information preservation and confusion suppression.
    \item Extensive experiments demonstrate that Dual-CARE consistently enhances incremental detection accuracy across various replay generators and alleviates the negative effects of domain overlap.
\end{itemize}

\section{Motivation}\label{sec:moti}
As illustrated in Fig. \ref{fig:dist-intro}, when the replay generator differs from the forgery generator, the generation artifacts shared by the replayed real and fake samples remain comparable and therefore do not interfere with the detector’s learning. In contrast, when the replay generator closely resembles the forgery generator, its own generative artifacts are likely to be interpreted as forgery cues. This misalignment causes the replayed real samples to drift away from the true real distribution, thereby confusing the detector. We refer to this phenomenon as the \textbf{Domain Confusion Effect}.

Upon closer examination, this effect indicates that the replayed fake samples themselves do not substantially affect detection performance, regardless of the replay generator used. The core issue instead lies in the replayed real samples, which cannot always be safely used for training. Although their FID suggests that they approximate the original real distribution, these samples still contain subtle synthetic artifacts that can distort the classifier’s decision boundary. Therefore, the key challenge is to leverage the informative content embedded in these replayed real samples while mitigating their generative bias. Moreover, when the replay generator is dissimilar to the original forgery generator, the replayed real samples tend to be less disruptive and can be directly used for classifier training. Consequently, developing a unified learning strategy that remains effective across different replay generators becomes essential for robust generative replay in forgery detection.
\section{Methodology}
\begin{figure*}[t!]
    \centering
    \includegraphics[width=1\linewidth]{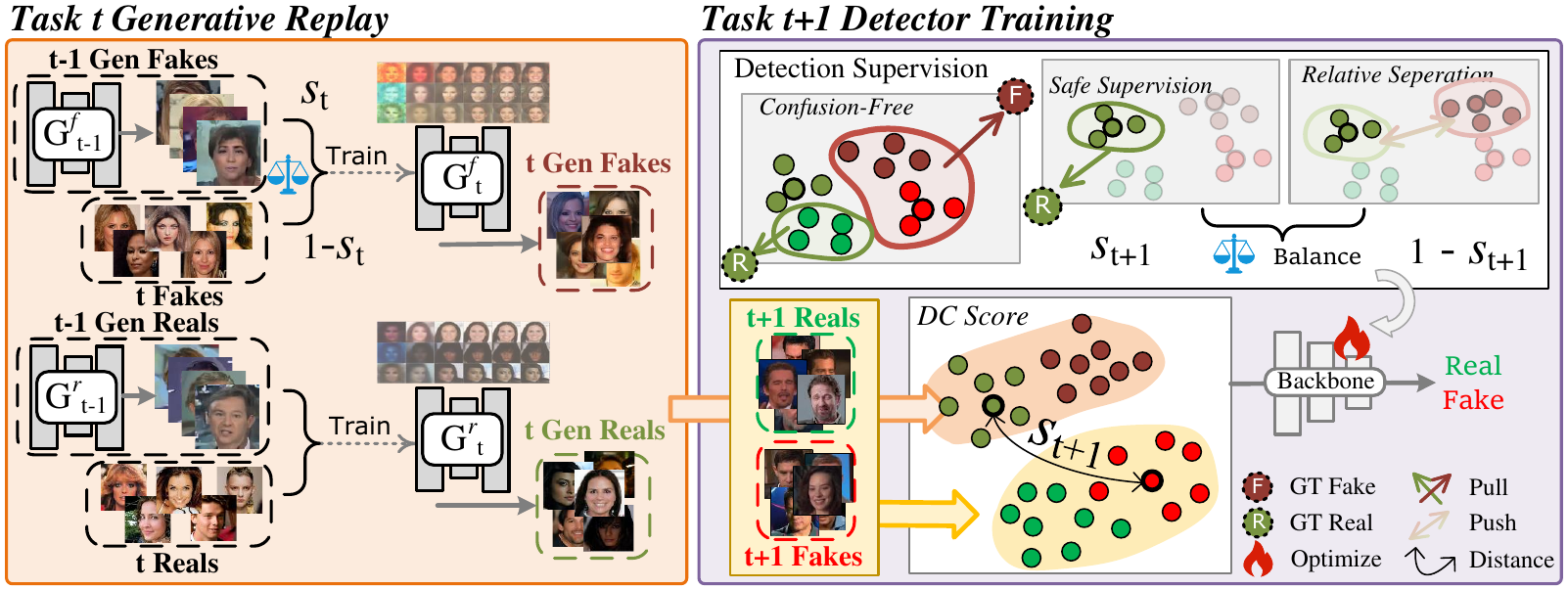}
    \caption{Overall framework of our method. We propose a dual confusion-aware regularization strategy to formulate the training of both generation and detection, thereby enabling the integration of generative replay into generative deepfake detection.}
    \label{fig:main-arch}
\end{figure*}
\subsection{Overall Framework}
In this paper, we propose a Dual \underline{C}onfusion-\underline{A}ware \underline{RE}gularization (Dual-CARE) strategy to fully exploit the potential of generative replay for generated face forgery detection. Specifically, we first introduce the Domain-aware Confusion Score (DC Score) to quantify the current condition of the domain confusion issue, thus \textbf{dual-modulating} the loss tradeoff of both generator and detector training. Then, we train a pair of updating replay generators to generate replay data that simulates all previous-task domains, with a domain-aware fast training strategy that leverages DC Score to enhance generative performance. Finally, we deploy the ``domain-safe'' samples to directly supervise the incremental deepfake detector, and we introduce Relative Separation Loss (RS Loss) as a tradeoff with the direct supervision for the ``domain-risky'' samples. The overall framework of our method is shown in Fig.~\ref{fig:main-arch}.

\subsection{Domain-aware Confusion Score}
Here, we design a Domain-aware Confusion Score (DC Score) to quantify the domain confusion between current generative forgery data and the deployed replay generator. Since a distinct boundary between domain-risky and domain-safe data is natively absent, the DC Score adaptively measures this confusion to dynamically balance the weights of direct and relative constraints. Formally, given a group of images $\mathbf{X}$, their centroid is defined as:
\begin{equation}
Cent(\mathbf{X}) = \frac{1}{|\mathbf{X}|} \sum_{\mathbf{x} \in \mathbf{X}} f(\mathbf{x}), \label{eq:center}
\end{equation}
where $f(\cdot)$ is the feature extractor. Following Eq.~\ref{eq:center}, let $Cent(\mathcal{D}^{r\text{-gen}}_t)$ and $Cent(\mathcal{D}_{t+1}^{f})$ be the centroids of the historical generated real samples $\mathcal{D}^{r\text{-gen}}_t$ and the current fake data $\mathcal{D}_{t+1}^{f}$, respectively. Their distance is then quantified via $L_2$ distance as:
\begin{equation}     
    s_{t+1}= \|Cent(\mathcal{D}^{r\text{-gen}}_t) - Cent(\mathcal{D}_{t+1}^{f})\|_2. 
    \label{eq:dcs_dist} 
\end{equation} 
The normalized $s_{t+1}$ is employed as the DC score to directly measure distribution separation, where a smaller value indicates a higher confusion risk.

\subsection{Diffusion Replay Generation for IFFD}

\paragraph{\textbf{General Training Pipeline of Diffusion Generator.}}
Given a set of deepfake detection training data as $\mathcal{D}=\{\mathcal{D}^{f},\mathcal{D}^{r}\}$, we deploy Latent Diffusion Model (LDM) as our generator backbone considering the superior performance of the advanced diffusion model. Because $\mathcal{D}^{f}$ and $\mathcal{D}^{r}$ provide distinct optimization trajectories for the detector, we deploy two separate generators to model each domain respectively. Learning a single domain is defined as:
\begin{equation}
\mathcal{L}_{G^{(\cdot)}}(\mathcal{D}^{(\cdot)}) = 
\mathbb{E}
\left[
\left\| 
\epsilon - \epsilon_{\theta}\left(\mathbf{x}_s,\, s,\, \mathbf{c}\right)
\right\|_2^2
\right], \label{eq:gen}
\end{equation}
where ${\mathbf{x}\sim \mathcal{D}^{(\cdot)},\, \epsilon\sim\mathcal{N}(0,1),\, s\sim \mathcal{U}(1,S)}$. Based on Eq.~\ref{eq:gen}, we can obtain a pair of trained generators $\mathbf{G}=\{G^f,G^r\}$. 

\paragraph{\textbf{Generative Replay for All Previous Tasks in Deepfake Detection.}}
Here, we aim to obtain replay samples $\mathbf{R}^{\mathrm{gen}}_t$ for incremental learning at task $(t+1)$ through diffusion-based generative replay. With respect to the number of tasks, we maintain two jointly updated generators $\mathbf{G}_t=\{G_t^f,G_t^r\}$ to reproduce the historical fake and real distributions observed up to task $t$. In practice, the two generators are implemented using a fixed diffusion backbone with two lightweight LoRA adapters for the fake and real domains, respectively. Only the latest pair of adapters is retained after each incremental task, avoiding the need to store either historical raw samples or an increasing number of task-specific generators.

Specifically, following DDGR~\cite{gao2023ddgr}, we use the generators $\mathbf{G}_{t-1}$ obtained after task $(t-1)$ to synthesize a replay dataset $\mathcal{D}_{t-1}^{\mathrm{gen}}$, which recursively represents the distributions of all tasks up to task $(t-1)$. The generated replay data are subsequently combined with the current-task dataset $\mathcal{D}_t$ to update $\mathbf{G}_t$. Formally, the training data for each generator can be written as:
\begin{equation}
    \hat{\mathcal{D}}_t^{(\cdot)}
    =
    \{
    \mathcal{D}_t^{(\cdot)},
    \mathcal{D}_{t-1}^{(\cdot)\text{-gen}}
    \},
\end{equation}
where $(\cdot)\in\{r,f\}$ denotes the real or fake domain. In this manner, $\mathbf{G}_t$ learns the current-task distribution while retaining the generative knowledge accumulated from previous tasks. See Sec.~\ref{sec:computation} for detailed computation analysis.
\paragraph{\textbf{Domain-aware Fast Training.}}
Since generation performance natively depends on the targeted distribution characteristics, we deploy a confusion coefficient to regularize the generator's training process. Specifically, we first decouple the training gradients induced by the previous and current data, which yields:
$\mathcal{L}_{G^{(\cdot)}}(\mathcal{\hat{D}}_t^{(\cdot)})=\mathcal{L}_{G^{(\cdot)}}(\mathcal{D}_t^{(\cdot)})+\mathcal{L}_{G^{(\cdot)}}(\mathcal{D}_{t-1}^{(\cdot)\text{-gen}})$.
For the real generator $G^{r}$, our objective is to approximate multiple real domains as faithfully as possible. The real domains and the synthetic artifacts introduced by the generator itself are not inherently conflicting during the generation process. The potential conflict with fake domains arises primarily during detection, which should therefore be addressed through appropriate design at the detector training stage. Consequently, $G^{r}$ is expected to be effective in the acquisition of new knowledge while preserving previously learned knowledge. Hence, no additional weighting mechanism is required for $G^{r}$. 

However, samples generated by $G^{f}$ inherently carry intrinsic fabrication artifacts, complicating their relationship with newly learned forgery features. To stabilize training and enhance performance, we introduce a confusion coefficient to adaptively balance the training weights:
\begin{equation}
  \mathcal{L}_{G^{f}}(\mathcal{\hat{D}}_t^{f})=(1-s_{t})\times\mathcal{L}_{G^{f}}(\mathcal{D}_t^{f})+s_{t}\times\mathcal{L}_{G^{f}}(\mathcal{D}_{t-1}^{f\text{-gen}}).  
\end{equation}
As $s_t$ increases, the domain divergence between current-task forgery artifacts and the generator's inherent characteristics becomes increasingly pronounced. In such cases, the gradient conflict introduced by incoming information aggravates, rendering the preservation of previously acquired knowledge paramount to prevent catastrophic overwriting. Accordingly, increasing the penalty weight for prior knowledge facilitates a gradual adaptation process, thereby safeguarding historical performance and guiding the model toward a more favorable local optimum. Conversely, when $s_t$ is small, the domain discrepancy between the current task and the generator is bounded, thereby alleviating the conflicts induced by incoming knowledge. Under this condition, a larger learning weight can be assigned to the new information gradient, which facilitates faster convergence while maintaining stability. 


\paragraph{\textbf{Final Summarized Samples for Detector Training.}}
During incremental learning of deepfake detectors,  $\mathbf{G}$ will provide the generative replays $\mathbf{R}^{\text{gen}}_t=\{\mathcal{D}_{t}^{f\text{-gen}},\mathcal{D}_{t}^{r\text{-gen}}\}$, which will be combined with the next ($t+1$)-th data for training. During ($t+1$)-th task, the training mini-batch is the combination of generative real/fake replay and ($t+1$)-th data, which can be written as $\mathbf{X}_{t+1}=\{\mathbf{X}^r_{t+1},\mathbf{X}^f_{t+1},\mathbf{X}_g^r,\mathbf{X}_g^f\}$. Here, each $\mathbf{X}^{(\cdot)}_{(\cdot)}$ is a group of corresponding $\mathbf{x}^{(\cdot)}_{(\cdot)}$, while $\{\mathbf{X}^r_{t+1},\mathbf{X}^f_{t+1}\}\subseteq\mathcal{D}_{t+1}$ and $\{\mathbf{X}_g^r,\mathbf{X}_g^f\} \subseteq \mathbf{R}^{\text{gen}}_t$.

\begin{table*}[t]
\centering
\caption{Quantitative comparison (AUC) on Protocol 1 (Mixed-Era Forgery Incremental). L-Bound denotes vanilla incremental learning. Pre Avg. denotes the average performance over previous tasks. AF $\downarrow$ denotes Average Forgetting.}
\footnotesize
\resizebox{\linewidth}{!}{
\begin{tabular}{l @{\hspace{3pt}}c c  |c c c c c c  |c c c}
\toprule
Method & Venue & Task & LDM & DFDCP & SDv21 & DDPM & DiT & CDF & Pre Avg. & AF $\downarrow$ & PD $\downarrow$ \\
\toprule

\multirow{6}{*}{L-Bound} & \multirow{6}{*}{--} & T1 & 99.99 & - & - & - & - & - & - & - & - \\
& & T2 & 80.75 & 88.76 & - & - & - & - & 80.75 & 19.24 & 15.23 \\
& & T3 & 77.46 & 52.63 & \textbf{99.99} & - & - & - & 65.04 & 29.33 & 23.30 \\
& & T4 & 35.18 & 55.52 & 31.75 & 99.93 & - & - & 40.81 & 55.43 & 44.39 \\
& & T5 & 51.44 & 50.77 & 70.63 & 77.10 & 99.21 & - & 62.48 & 34.68 & 30.16 \\
& & T6 & 52.73 & 80.54 & 81.78 & 50.34 & 53.30 & \textbf{99.56} & 63.73 & 33.84 & 30.28 \\
\midrule

LwF & TPAMI' 17 & T6 & 81.31 & 78.16 & 89.40 & 49.11 & 76.92 & 98.15 & 74.98 & 34.51& 21.15 \\
iCaRL & CVPR' 17 & T6 & 83.72 & 76.31 & 84.54 & 59.75 & 71.91 & 99.20 & 75.25 & 19.45& 18.60 \\
DER & CVPR' 21 & T6 & 77.31 & 80.12 & 90.36 & 82.01 & 83.23 & 99.07 & 82.61 & 30.08& 15.75 \\
CoReD & MM' 21 & T6 & 88.58 & 83.79 & 94.97 & 78.79 & 71.32 & 99.17 & 83.49 & 6.06& 17.52 \\
HDP & IJCV' 24 & T6 & 94.31 & 85.82 & 93.73 & 95.10 & 81.53 & 99.40 & 90.10 & 4.63& 16.03 \\
\midrule

\multirow{6}{*}{DFIL} & \multirow{6}{*}{MM' 23} & T1 & 99.99 & - & - & - & - & - & - & - & - \\
& & T2 & 97.22 & 91.07 & - & - & - & - & 97.22 & 2.77 & 5.85 \\
& & T3 & 98.23 & 75.91 & 99.98 & - & - & - & 87.07 & 8.46 & 8.62 \\
& & T4 & 94.01 & 72.43 & 99.80 & \textbf{99.98} & - & - & 88.75 & 8.27 & 8.43 \\
& & T5 & 95.84 & 66.56 & 98.61 & 88.32 & 96.31 & - & 87.33 & 10.42 & 10.86 \\
& & T6 & 92.37 & 73.39 & 99.02 & 84.27 & 77.39 & 99.48 & 85.29 & 12.18 & 12.33 \\
\midrule

\multirow{6}{*}{SUR-LID} & \multirow{6}{*}{CVPR' 25} & T1 & 99.99 & - & - & - & - & - & - & - & - \\
& & T2 & 99.31 & 90.32 & - & - & - & - & \textbf{99.31} & 0.68 & 5.17 \\
& & T3 & 99.17 & 87.95 & 99.96 & - & - & - & 93.56 & 1.59 & 4.30 \\
& & T4 & \textbf{99.50} & 80.27 & 99.54 & 99.73 & - & - & 93.10 & 3.65 & 5.23 \\
& & T5 & 98.70 & 75.11 & \textbf{99.62} & \textbf{99.89} & \textbf{99.55} & - & 93.33 & 4.17 & 5.41 \\
& & T6 & 97.40 & 82.18 & \textbf{99.03} & \textbf{98.35} & 92.50 & 99.22 & 93.89 & 4.02 & 5.21 \\
\midrule

\multirow{6}{*}{\textbf{Dual-CARE (Ours)}} & \multirow{6}{*}{--} & T1 & 99.99 & - & - & - & - & - & - & - & - \\
& & T2 & \textbf{99.67} & \textbf{91.50} & - & - & - & - & \textbf{99.67} & \textbf{0.32} & \textbf{4.41} \\
& & T3 & \textbf{99.60} & \textbf{91.78} & 99.97 & - & - & - & \textbf{95.69} & \textbf{0.05} & \textbf{2.87} \\
& & T4 & 93.63 & \textbf{90.91} & 99.25 & \textbf{99.98} & - & - & \textbf{94.60} & \textbf{2.56} & \textbf{4.05} \\
& & T5 & \textbf{99.08} & \textbf{86.96} & \textbf{99.62} & 96.45 & 95.59 & - & \textbf{95.53} & \textbf{2.33} & \textbf{4.45} \\
& & T6 & \textbf{98.45} & \textbf{87.36} & 97.71 & 95.28 & \textbf{98.07} & 98.93 & \textbf{95.37} & \textbf{2.03} & \textbf{4.03} \\

\bottomrule
\end{tabular}}
\label{tab:main_comparison}
\end{table*}
\subsection{IFFD Training with Generative Replay}
\paragraph{\textbf{Relative Separation Loss.}}
As previously discussed, domain confusion among generative real replays, actual real samples, and actual fake samples may mislead the detector when direct label-based supervision is conducted. However, the similar learning effectiveness between domain-safe\&real and domain-risky\&real suggests that the domain-risky samples also contain previous information that could be beneficial to mitigating the catastrophic forgetting. Therefore, we propose the Relative Separation Loss (RS Loss) to leverage the valuable previous information. Instead of explicit real-fake supervisions, RS Loss considers from an indirect relative perspective. Specifically, despite the domain-risky $\mathbf{x}_g^r$ may exhibit similarity with current fake samples, its relative relation with $\mathbf{x}_g^f$ should be consistent. This is because both $\mathbf{x}_g^r$ and $\mathbf{x}_g^f$ are generated by the same real domain and the same generative character. As a result, the distinction between them is effectively nullified and filtered, leaving only the $t$-th forgery character. Consequently, the distribution of $\mathbf{x}_g^f$ and $\mathbf{x}_g^r$ could be encouraged to separate if they correspond to domain-risky samples. 

To be specific, given a minibatch $\mathbf{X}$, we first calculate the feature centroid of the generated real samples $\mathbf{X}^r_g$ based on Eq.~\ref{eq:center}, denoted as $Cent(\mathbf{X}^r_g)$.
Then, we define RS Loss to maximize the separation between this generated real centroid and each sample-wise generated fake feature $\mathbf{x} \in \mathbf{X}^f_g$. This is achieved by minimizing their average cosine similarity, which can be written as:
\begin{equation}
\mathcal{L}_\text{rs} = \frac{1}{|\mathbf{X}^f_g|} \sum_{\mathbf{x} \in \mathbf{X}^f_g} \frac{f(\mathbf{x}) \cdot Cent(\mathbf{X}^r_g)}{\|f(\mathbf{x})\|_2 \cdot \|Cent(\mathbf{X}^r_g)\|_2}. \label{eq:rs_loss}
\end{equation}
RS Loss can leverage the previous information in the replay real sample by sample. By considering the relative authenticity differences among generated images, we indirectly exploit the information embedded within the generative replay, even when they are domain risky.
\paragraph{\textbf{Overall Detection Loss.}}
When domain confusion is relatively mild, direct supervision evidently provides strong guidance. Meanwhile, RS Loss facilitates maximal exploitation of the informative content within domain-risky data. Therefore, we propose to adaptively adjust the balance between RS Loss and direct supervision. 
Hence, by incorporating RS Loss and DC Score, we formulate a unified loss function that simultaneously optimizes both domain-risky and domain-safe samples for generative replay-based deepfake detection.
Firstly, the common supervision signal for face forgery detection is a Cross-Entropy loss, which could be written as:
\begin{equation}
        \mathcal{L}_{\text{ce}}(\mathbf{x}) = - \left( \hat{y} \log(y_p) + (1 - \hat{y}) \log(1 - y_p) \right),\label{eq:celoss}
\end{equation}
where $\hat{y}$ is the corresponding ground-truth label, $y_p$ is the predicted result from the backbone. Since $\forall\,\mathbf{x} \in \left\{ \mathbf{X}^r_{t+1},\ \mathbf{X}^f_{t+1},\ \mathbf{X}_g^f \right\}$ $\mathbf{x}$ is \textit{confusion-free} based on the discussion in Sec.~\ref{sec:moti}, we can directly constrained them by $\mathcal{L}_{cf}=\mathcal{L}_{ce}(\mathbf{x})$. Then, we separate $\tilde{\mathbf{x}} \in \mathbf{X}_g^r$ from $\mathbf{X}$ that could \textit{confuse} to adaptively adjust the direct and relative constraints based on DC Score, which can be written as:
\begin{equation}
     \mathcal{L}_{{c}}=s_{t+1} \mathcal{L}_{ce}(\tilde{\mathbf{x}})+ (1-s_{t+1}) \mathcal{L}_{rs},
\end{equation}
which means the safe direct supervision could be conducted with higher weights if the domain distance is relatively large, otherwise a higher $L_{rs}$ should be applied.

Therefore, the overall loss can be written as:
\begin{equation}
    \mathcal{L}_{overall}=\mathcal{L}_{c}+\mathcal{L}_{cf},
\end{equation}
which encourages the detector to leverage information from current and replay samples simultaneously.


\section{Experimental Results}

\subsection{Experimental Settings}
\begin{figure}[t]
\centering
\begin{minipage}{0.3\textwidth}
  \centering
  \includegraphics[width=\linewidth]{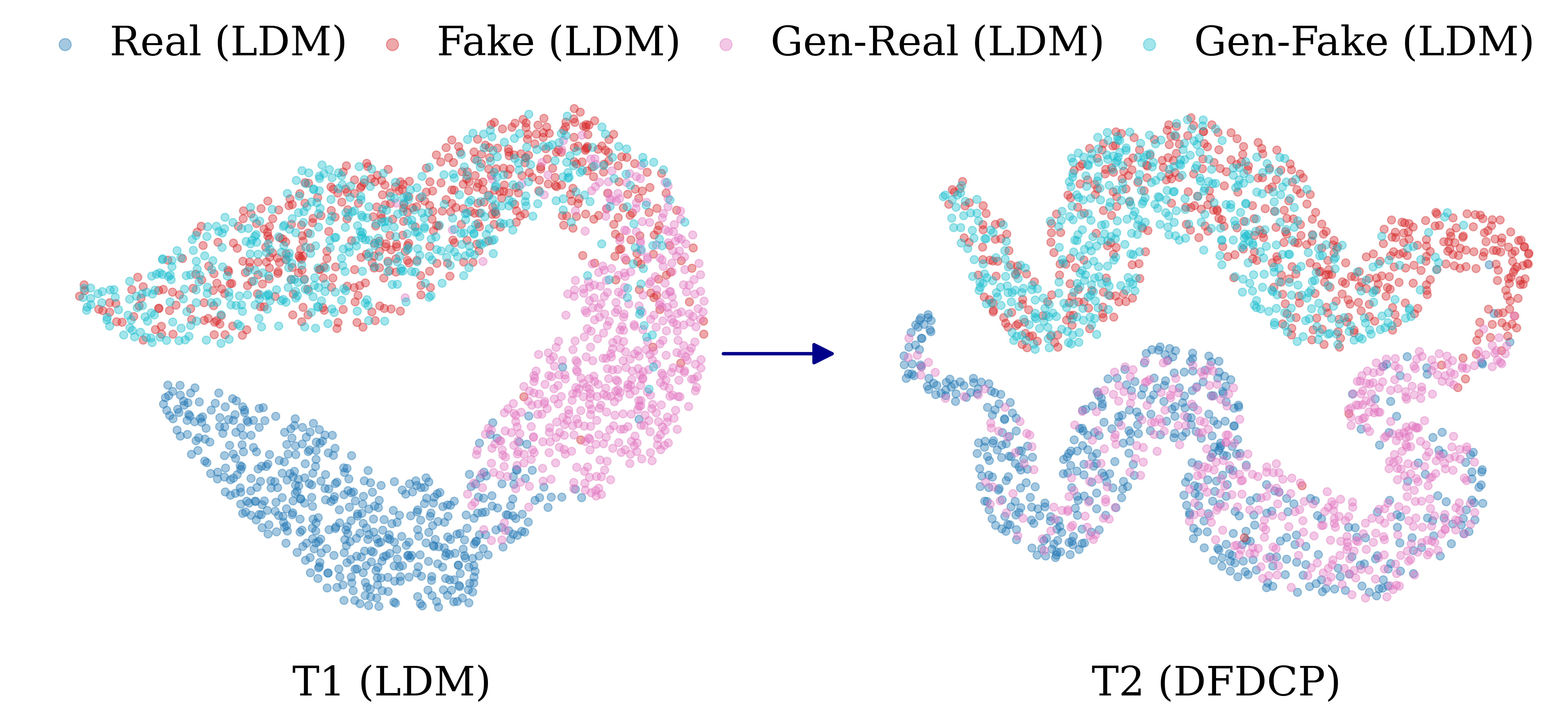}
  \caption{UMAP visualizations. (Left) Initial state: Gen-Real (pink) is ambiguously positioned near Fake (red / cyan). (Right) After learning T2: Our method ($\mathcal{L}_{rs}$) actively separates the Gen-Real cluster, resolving the domain confusion.}\label{fig:feature}
\end{minipage}
\hfill
\begin{minipage}{0.16\textwidth}
  \centering
  \includegraphics[width=\linewidth]{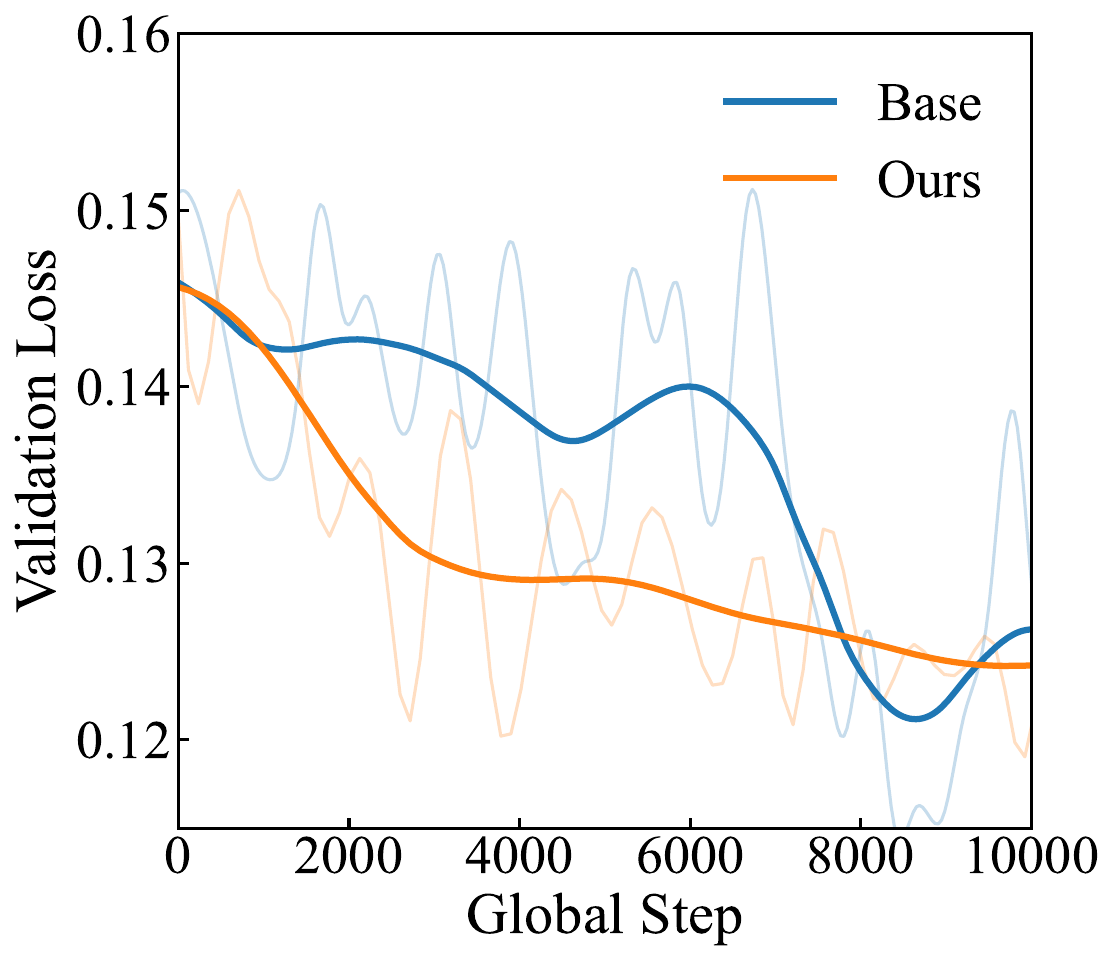}
  \caption{Learning performance with (Ours) or without (Base) Domain-aware Fast Training.}\label{fig:loss}
\end{minipage}
\end{figure}
\begin{figure}[t]
    \centering
    \includegraphics[width=1\linewidth]{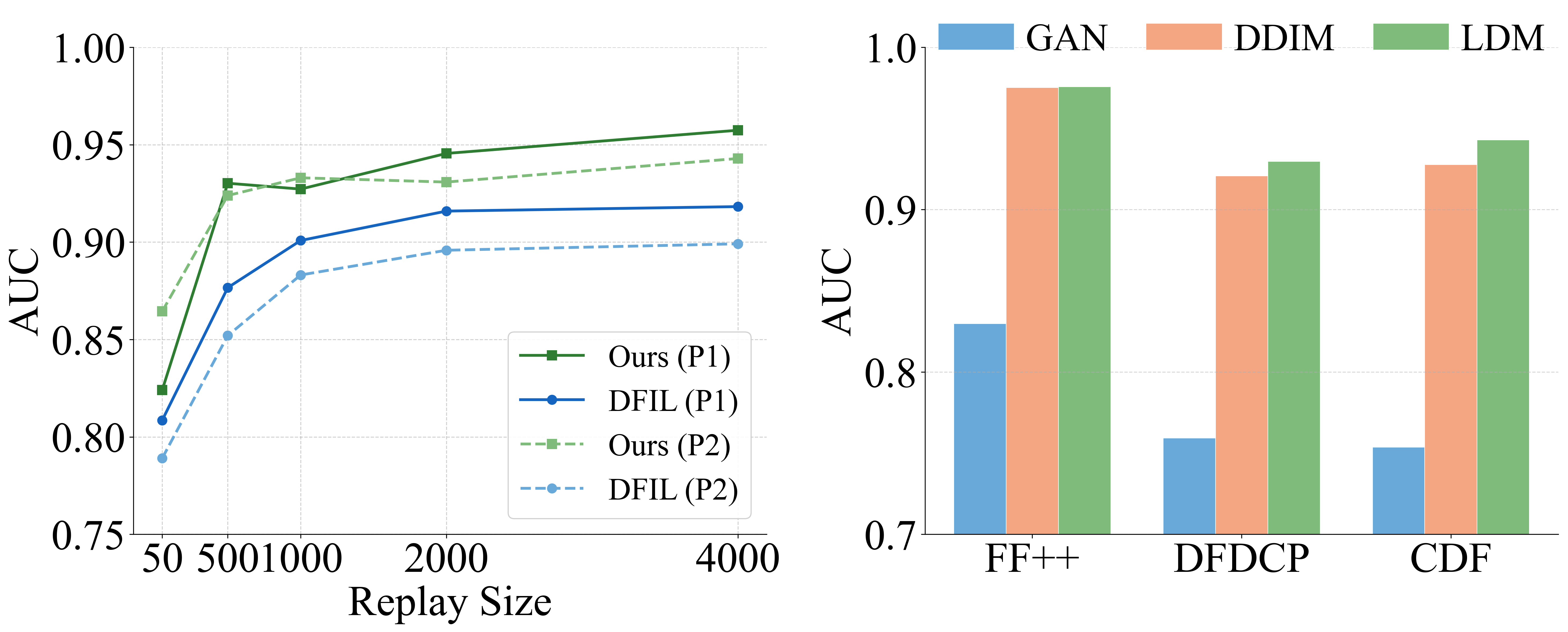} 
    \caption{Analysis of generative replay strategy. (Left) Performance vs. replay sample size on P1 and P2. (Right) Impact of replay generator quality (avg. AUC on T2-T4).}
    \label{fig:replay_analysis}
\end{figure}
\subsubsection{Datasets.}
To construct a comprehensive and challenging benchmark for incremental face forgery detection, our experiments utilize a curated selection of datasets, which is designed to simulate a realistic scenario by spanning both classical, widely-used forgery datasets and the latest cutting-edge forgeries generated by advanced diffusion models. The classical datasets include: Celeb-DF-v2 (CDF)~\cite{Celeb-df}, DeepFake Detection Challenge Preview (DFDCP)~\cite{dfdc}, and the hybrid-category FaceForensics++ (FF++)~\cite{FF++}. To address the most recent threats, we further incorporate a suite of modern forgeries, that is, \{SDv21~\cite{sdv15}, DiT~\cite{DiT}\} from DF40~\cite{df40} and \{LDM~\cite{rombach2022high}, DDPM~\cite{ho2020denoising}\} from DiffusionFace~\cite{diffusionface}. This blend of classical and cutting-edge forgery types creates a comprehensive evaluation against evolving threats.


\subsubsection{Incremental Protocols.}
To comprehensively evaluate model robustness in evolving forgery landscapes, we propose two complementary incremental protocols that capture real-world dynamics and benchmark-level comparability.


\begin{itemize}

\item \textbf{Protocol 1 (P1): Mixed-Era Forgery Incremental.}
It follows the sequence \{LDM, DFDCP, SDv21, DDPM, DiT, CDF\}, simulating a realistic, temporally chaotic evolution of forgery techniques. It intentionally interleaves classical Face-Swapping (FS) datasets (DFDCP, CDF) with modern Entire Face Synthesis (EFS) forgeries to emulate the heterogeneous and non-sequential emergence of threats in the wild. This setup is designed to rigorously evaluate the resilience of a model to catastrophic forgetting and its adaptability to domain confusion.


\item \textbf{Protocol 2 (P2): Benchmark-Aligned Incremental.}  
It employs the sequence \{DDPM, FF++, DFDCP, CDF\} and extends the benchmark protocol introduced in recent work SUR-LID. To align with prior baselines while avoiding redundant configurations, we replace the initial dataset (SDv21) with DDPM, a diffusion-based forgery type. This modification ensures consistency with established benchmarks while incorporating emerging generative paradigms for a fair yet forward-looking evaluation.

\end{itemize}


\subsubsection{Implementation Details.}
Our framework utilizes an EfficientNetB4~\cite{effnet} backbone, trained via the Adam optimizer~\cite{adam} with a learning rate of 0.0002 for 5 epochs. Inputs are resized to $256 \times 256$ with a batch size of 32. For sample-replay baselines, the buffer size is 500 samples per task. All baselines are replicated within DeepFakeBench~\cite{deepfakebench} to ensure fair comparison. We adopt frame-level AUC~\cite{deepfakebench} as the primary metric, supplemented by accuracy (ACC). To quantify catastrophic forgetting, the Performance Dropping rate is defined as $PD = M_0 - M_N$, where $M_0$ and $M_N$ denote the average metrics in the base and final sessions, respectively. All experiments are conducted on an NVIDIA A100 GPU.

\begin{table}[t] \centering \caption{\small Practical Cost Comparisons. $n$ denotes replay size.} \label{tab:time} \scriptsize \setlength{\tabcolsep}{2pt} \renewcommand{\arraystretch}{0.1} \begin{tabular}{lcccc} \toprule Method & GenTrain (h) & DetTrain (h) & Storage (MB) & Inference (s) \\ \midrule Common & -- & 1.31 & $0.3n$ & $8.03{\times}10^{-4}$ \\ \textbf{Ours (Gen)} & 1.06 & $1.31{+}1.2{\times}10^{-5}n$& 38.2 & $8.03{\times}10^{-4}$ \\ \bottomrule \end{tabular} \end{table}
\begin{table}[t]
\centering
\caption{\small Ablation study on core components and distance metrics.
All results are AUCs obtained by the final Protocol~1 model at T6
and evaluated on tasks T1--T5.
$\Delta_{\mathrm{Avg}}$ denotes the average performance difference
relative to Ours.
Within each group, the best and second-best results are highlighted
in \textbf{bold} and \underline{underline}.}
\label{tab:ablation}

\scriptsize
\setlength{\tabcolsep}{1.6pt}
\renewcommand{\arraystretch}{1.05}

\begin{tabular}{lccccccc}
\toprule
Variant
& LDM
& DFDCP
& SDv21
& DDPM
& DiT
& Avg.
& $\Delta_{\mathrm{Avg}}$ \\
\midrule

\multicolumn{8}{l}{\textbf{\textit{Core component ablation}}} \\

\textit{w/o} DC-guided Gen
& \underline{95.42}
& 83.61
& 94.16
& \underline{96.38}
& \textbf{94.99}
& \underline{92.91}
& $-2.07$ \\

\textit{w/o} Gen-Real Sup.
& 88.41
& 79.17
& \underline{98.52}
& 52.04
& 53.47
& 74.32
& $-20.66$ \\

\textit{w/o} $\mathcal{L}_{rs}$
& 92.75
& \textbf{89.29}
& \textbf{99.34}
& 90.47
& 83.04
& 90.97
& $-4.01$ \\

\textbf{\textit{w} All (Ours)}
& \textbf{99.69}
& \underline{89.07}
& 98.39
& \textbf{99.47}
& \underline{88.29}
& \textbf{94.98}
& -- \\

\midrule

\multicolumn{8}{l}{\textbf{\textit{Distance metric ablation}}} \\

DCS-Cos + RS-Cos
& 97.96
& \textbf{92.59}
& \underline{97.52}
& 98.20
& 74.02
& 92.05
& $-2.93$ \\

DCS-L2 + RS-L2
& \textbf{99.77}
& 85.62
& 96.09
& 97.12
& 79.10
& 91.54
& $-3.44$ \\

DCS-Cos + RS-L2
& 99.54
& 86.37
& 97.31
& \underline{99.28}
& \underline{82.05}
& \underline{92.91}
& $-2.07$ \\ 

\textbf{DCS-L2 + RS-Cos (Ours)}
& \underline{99.69}
& \underline{89.07}
& \textbf{98.39}
& \textbf{99.47}
& \textbf{88.29}
& \textbf{94.98}
& -- \\

\bottomrule
\end{tabular}
\end{table}

\begin{figure}[b]
    \centering
    \includegraphics[width=1\linewidth]{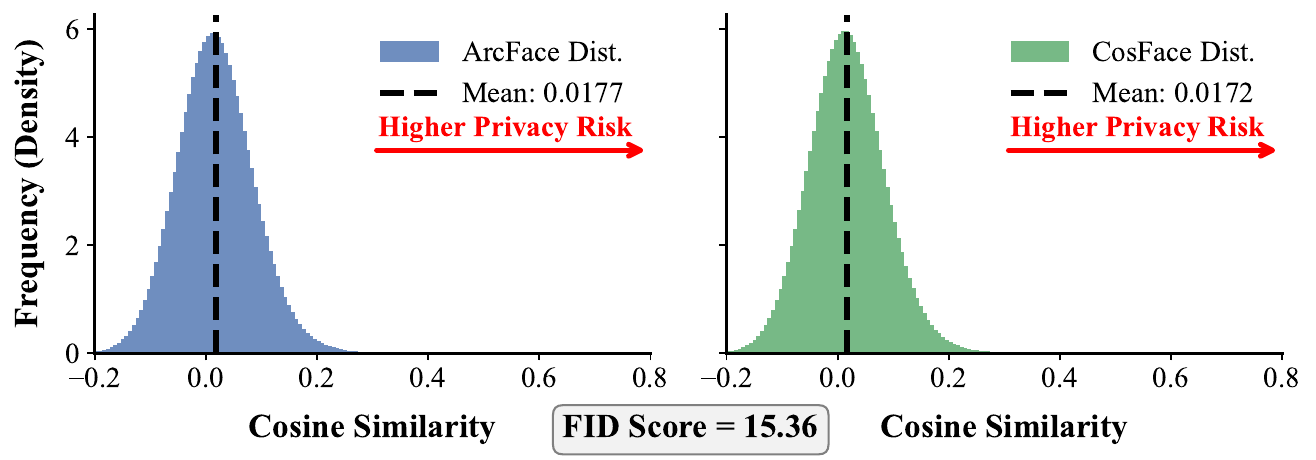}
    \caption{Quantitative analysis on privacy protection effectiveness of our method.} 
    \label{fig:id-leak}
\end{figure}

\begin{table*}[htbp]
\centering
\caption{Performance comparison of various generative replay strategies on Protocol 1. We evaluate Full Replay, Fake-Only Replay, our dynamic Dual-CARE, and two baselines with fixed confusion score.}
\footnotesize
\setlength{\tabcolsep}{6.5pt}
\begin{tabular}{l | c c c c c c c c | c c c c}
\toprule
\multirow{2}{*}{Method} & \multicolumn{8}{c|}{Task-Incremental Performance (AUC)} & \multicolumn{4}{c}{Average Accuracy (ACC)} \\

\cmidrule(lr){2-9} \cmidrule(lr){10-13}
& LDM & DFDCP & SDv21 & DDPM & DiT & CDF & Avg. & PD $\downarrow$ & Real & PD $\downarrow$ & Fake & PD $\downarrow$ \\
\toprule

Lower Bound & 52.73 & 80.54 & 81.78 & 50.34 & 53.30 & 99.56 & 69.71 & 30.28 & 89.51 & 10.34 & 41.28 & 58.62 \\
Full Replay & 96.59 & 84.61 & 94.96 & 98.08 & 75.67 & 99.37 & 91.55 & 8.44 & 84.55 & 15.15 & 83.96 & 15.89 \\
Fake-Only Replay & 95.39 & 76.24 & 98.09 & 49.94 & 52.10 & 99.63 & 78.56 & 21.43 & 82.77 & 16.98 & 63.10 & 36.80 \\
Fixed $s_t=0.5$ & 99.77 & 87.57 & 97.25 & 99.40 & 78.66 & 99.70 & 93.72 & 6.27 & 94.44 & 5.36 & 62.36 & 37.49 \\
Fixed $s_t=0.1$ & 99.59 & 88.71 & 96.69 & 98.86 & 75.19 & 99.69 & 93.12 & 6.87 & \textbf{95.38} & \textbf{4.52} & 61.39 & 38.51 \\
\textbf{Ours (Adaptive)} & 99.69 & 89.07 & 98.39 & 99.47 & 88.29 & 99.51 & \textbf{95.74} & \textbf{4.25} & 89.60 & 10.25 & \textbf{85.29} & \textbf{14.56} \\

\bottomrule
\end{tabular}
\label{tab:Generation analysis}
\end{table*}
\subsection{Effectiveness Comparisons with Existing Methods}
As shown in Tab.~\ref{tab:main_comparison}, our method consistently outperforms general continual learning (LwF~\cite{li2017learning}, iCaRL~\cite{rebuffi2017icarl}, DER~\cite{yan2021dynamically}) and state-of-the-art IFFD-specific baselines (CoReD~\cite{kim2021cored}, HDP~\cite{sun2025continual}, DFIL~\cite{pan2023dfil}, SUR-LID~\cite{cheng2025stacking}) under Protocol 1, achieving the highest average AUC with \textbf{no} data replay. This advantage stems from two aspects: (1) generative replay approximates historical distributions to yield diverse rehearsal samples without storing real images; (2) confusion-aware regularization distinguishes reliable from ambiguous signals, facilitating knowledge transfer while minimizing negative interference to enhance stability. Protocol 2 comparisons are provided in the \textit{Appendix}.
\subsection{Practical Cost of Generative Replay}\label{sec:computation}

We analyze the practical cost of generative replay from three perspectives, with the results reported in Tab.~\ref{tab:time}.\\
\textbf{1) Generator training.} Generative replay introduces additional generator training cost at each incremental task. However, this cost is incurred only during training and does not affect the detector's inference efficiency during deployment. Therefore, it represents a reasonable trade-off for achieving privacy-preserving replay and scalable replay diversity.\\
\textbf{2) IFFD replay overhead.} During IFFD training, the additional computational overhead of the proposed Dual-CARE mainly arises from replay-sample generation, which only involves generator inference and is therefore practically minor.\\
\textbf{3) Storage cost.} Rather than storing an increasing number of raw replay samples, Dual-CARE maintains only a lightweight LoRA module, which can generate arbitrarily many replay samples without increasing the fixed storage burden. In contrast, sample-based replay requires storage to grow with the number of tasks.

\subsection{Ablation Study}
Tab.~\ref{tab:ablation} validates the component performance of Dual-CARE, demonstrating the effectiveness of DC-guided generation. Training curves in Fig.~\ref{fig:loss} further confirm that our domain-aware strategy effectively accelerates convergence when $s_t$ is small, perfectly aligning with our theoretical design. To evaluate the detector training strategy as the core solution to domain confusion, we isolated its components by disabling any generator strategy. Removing Gen-Real supervision triggers a substantial performance drop, underscoring the necessity of explicit label guidance for generated samples. Similarly, omitting $\mathcal{L}_{rs}$ degrades performance significantly, confirming that direct supervision and relative separation are complementary in mitigating generative artifacts and maintaining decision boundaries. Moreover, metric analysis reveals that all distance combinations outperform variants without $\mathcal{L}_{rs}$, while DCS-L2 + RS-Cos achieves the best results, indicating that L2 distance captures domain confusion better for DCS and cosine similarity more effectively enforces feature separation in $\mathcal{L}_{rs}$. Furthermore, supplementary analyses covering detailed component-wise ablations, alternative backbone choices, different normalization functions for computing the adaptive weight $\alpha$, and sample-wise versus centroid-based variants of $\mathcal{L}_{rs}$ are detailed in the \textit{Appendix}.





\subsection{Analysis of Generative Replay}

\subsubsection{Advantage on Replay Diversity.}
Unlike common replay which requires proportional storage overhead to scale, generative replay inherently yields infinite images via a single generator. As illustrated in Fig.~\ref{fig:replay_analysis}, expanding the replay size consistently boosts the performance of both DFIL and our method, validating the critical impact of replay diversity. Furthermore, at an equivalent replay size, our framework demonstrates superior generative diversity across TCE, CLIP-distance, and LPIPS metrics, as detailed in \textit{Appendix}.


\subsubsection{Impact of Generator Quality.}
We analyze replay generator quality by comparing GAN, DDIM, and LDM under the P2 protocol (Fig.~\ref{fig:replay_analysis}, right). Results show a direct correlation between generation fidelity and detection accuracy. The outdated GAN architecture performs poorly, whereas diffusion-based LDM and DDIM generate promising replays that significantly enhance incremental learning, leading us to select LDM. Furthermore, the similar performance of LDM and DDIM proves that Dual-CARE is generator-agnostic (given sufficient generation capability), highlighting its scalability and extensive application potential.


    
    
\subsection{Visualization Analysis on Domain Confusion}
To intuitively analyze the observed domain confusion, we visualize the T1 LDM feature space using UMAP~\cite{umap}. As shown in Fig.~\ref{fig:feature}, the initial Gen-Real samples occupy an ambiguous region, lying close to the Fake clusters and thus presenting a high risk of domain confusion. In contrast, the right plot shows the feature space after incrementally learning T2 DFDCP. Guided by our adaptive mechanism driven by $\mathcal{L}_{rs}$, the ambiguity is effectively resolved: the Gen-Real features are pushed away from the Fake clusters and aligned with the Real cluster. This visualization shows that our method reliably rectifies distributional misalignment and restores clear feature separation.
Additional visualizations across multiple datasets are provided in the \textit{Appendix} to further support these observations. 

\subsection{Analysis of Domain Confusion Effect}
To analyze domain confusion, we visualize the T1 LDM feature space using UMAP~\cite{umap}. As illustrated in Fig.~\ref{fig:feature}, initial Gen-Real samples reside in an ambiguous region near Fake clusters, posing a high risk of confusion. Conversely, after incrementally learning T2 DFDCP, our $\mathcal{L}_{rs}$-driven adaptive mechanism effectively resolves this ambiguity, pushing Gen-Real features away from Fake clusters and aligning them with the Real cluster. This confirms that our method rectifies distributional misalignment and restores clear feature separation. Additional cross-dataset visualizations are provided in the \textit{Appendix}.

\subsection{Privacy Quantitative Analysis}
Traditional sample replay poses privacy risks by storing raw facial images. Generative replay avoids this but may risk identity memorization. To evaluate this, we measure feature similarity between synthesized replay samples and original training faces using ArcFace \cite{deng2018arcface} and CosFace \cite{wang2018cosface}. As shown in Fig.~\ref{fig:id-leak}, similarities are near zero with minimal variance. Despite a high generation fidelity (FID 15.36), the generator produces entirely novel identities by capturing the structural distribution rather than memorizing training samples. This confirms that our generative replay framework effectively mitigates identity leakage, providing a secure rehearsal mechanism for incremental deepfake detection.
Furthermore, in Appendix Tab.~\ref{tab:supp-privacy}, we demonstrate that the generative ID cannot be effectively recalled by the training samples. And GenReplay cannot be distinguished from the Actual Replay.


\section{Conclusion}
In this paper, we present Dual Confusion-Aware REgularization (Dual-CARE), a framework for enhancing generative replay in incremental face forgery detection. We analyze domain confusion between replay generators and new forgery models, identifying domain-safe samples for direct supervision and domain-risky samples that require adaptive handling. Guided by a Domain Confusion Score (DC Score), Dual-CARE dual-modulates generator updates and detector supervision, applying a Relative Separation Loss to domain-risky samples to balance informative cues and potential confusion. Extensive experiments show that Dual-CARE improves robustness and accuracy across diverse generative replay settings, while mitigating the negative impact of domain overlap, validating its effectiveness for incremental learning under evolving forgery threats.


\bibliography{AAAI2027/AnonymousSubmission2027}

\clearpage
\clearpage
\section*{Appendix}


\setcounter{section}{0}

\section{Related Works}
\label{sec:related_work}

\subsection{Face Forgery Detection}
Current face forgery detection methods typically leverage available forgery samples to train a generalized model capable of handling unseen forgeries. Various forgery-specific patterns, such as noise \cite{li2020face}, local region \cite{chen2021local,zhao2021multi}, and frequency information \cite{qian2020thinking,guo2020learning,kashiani2025freqdebias}, are explored to capture more discriminative forgery cues. To alleviate the performance degradation observed in cross-domain evaluations, researchers propose a range of learning strategies from different perspectives, including contrastive learning \cite{sun2022dual}, identity information modeling \cite{huang2023implicit,dong2023implicit}, disentangled representation learning \cite{liang2022exploring,yan2023ucf}, reconstruction-based learning \cite{cao2022end,wang2021representative}, and data augmentation \cite{chen2022self,shiohara2022detecting,yan2024transcending}. Recently, several ViT-based methods such as CLIP \cite{cui2025forensics} and LoRA-based Effort \cite{yan2025effort} are proposed to enhance the generalization capability of forgery detection by leveraging large vision-language models. In summary, many general approaches have been proposed to learn transferable forgery features from limited known data. These methods aim to maintain good performance on unseen samples. However, given the large scale and diversity of existing forgery data, relying on a few known datasets to train a truly universal detector is unrealistic.

\subsection{Incremental Learning for Forgery Detection}
Incremental learning has been extensively studied across various domains and is typically categorized into parameter isolation \cite{de2021continual}, parameter regularization \cite{aljundi2018memory,kirkpatrick2017overcoming,li2017learning}, and data replay \cite{mai2021supervised,rebuffi2017icarl}. In the field of face forgery detection, most incremental methods are based on sample replay, where representative samples from previous tasks are stored or reused to mitigate catastrophic forgetting. Representative replay-based approaches in incremental face forgery detection have adopted different strategies to preserve prior knowledge. For example, CoReD \cite{kim2021cored} relies on distillation loss to maintain knowledge from previous tasks. Meanwhile, DFIL \cite{pan2023dfil} improves replay effectiveness by emphasizing both center and hard samples. In addition, HDP \cite{sun2025continual} employs refined universal adversarial perturbations as a replay mechanism. Similarly, DMP \cite{tian2024dynamic} constructs mixed prototypes to summarize earlier task distributions. More recently, SUR-LID \cite{cheng2025stacking} introduces sparse uniform replay combined with a latent-space incremental detector to better preserve previous knowledge. Although effective in retaining previous knowledge, all these methods depend on explicit access to stored samples, which may raise privacy concerns and limit scalability.

Beyond sample-based replay, generative replay offers a promising alternative, reconstructing past distributions through generation rather than storage. This idea has been widely explored in general continual learning, where generative models synthesize pseudo-data to approximate previous task distributions and thus preserve earlier knowledge without explicit memory buffers. The seminal work Deep Generative Replay \cite{shin2017continual} introduces a dual-model framework that reconstructs past data distributions without storing real samples. Building on this idea, data-free class-incremental learning methods \cite{smith2021always} synthesize pseudo-samples through model inversion to enable continual learning under memory and privacy constraints. More recently, diffusion-based generative replay approaches such as DDGR \cite{gao2023ddgr} and SDDGR \cite{kim2024sddgr} have improved the stability and diversity of generated data, achieving stronger knowledge retention in incremental classification and detection tasks. These advances further demonstrate the substantial and growing potential of generative replay in continual visual learning, though its application to forgery detection remains largely unexplored.

\section{Detailed Implementation Settings}

\subsection{Preprocessing}

\paragraph{Data Preprocessing.}
Following the standard DeepFakeBench~\cite{deepfakebench} protocol, all video frames undergo face detection, extraction, and alignment before being resized to $256 \times 256$. For input normalization, we adopt a mean of $[0.5, 0.5, 0.5]$ and a standard deviation of $[0.5, 0.5, 0.5]$ for the three RGB channels. During training, 8 frames are uniformly sampled from each video, while 32 frames are sampled during testing to ensure more stable and reliable performance evaluation.

\paragraph{Data Augmentation.}
To enhance the generalization ability of the detector, we employ a comprehensive data augmentation pipeline on the current task data using the \texttt{Albumentations~\cite{buslaev2020albumentations}} library. The augmentation operations and their corresponding application probabilities are detailed as follows:


\begin{itemize}
\item \textbf{Spatial Transformations:} Horizontal Flip ($p=0.5$), Rotation within $\pm 10^\circ$ ($p=0.5$), and isotropic resizing.
\item \textbf{Pixel-level Transformations:} Gaussian Blur with a kernel size in the range of $[3, 7]$ ($p=0.5$).
\item \textbf{Compression Artifacts:} JPEG compression with a quality range of 40--100, applied with $p=0.5$.
\item \textbf{Color Perturbations:} One of Random Brightness/Contrast (limit 0.1), FancyPCA, or HueSaturationValue, selected and applied with probability $p=0.5$.
\end{itemize}

It is worth noting that while the current task data are augmented as described above, the generated replay samples are kept with standard normalization only so as to preserve their intrinsic generative distribution characteristics.

\subsection{Hyperparameters for Generative Replay}

In our generative replay setup, a replay buffer is constructed for each training batch. Specifically, for a current-task batch of size $B_{\text{new}} = 32$, we generate and replay $B_{\text{fake}} = 12$ fake samples and $B_{\text{real}} = 12$ real samples to maintain class balance during the incremental training process.

\subsection{Generative Model Details}

We adopt the Latent Diffusion Model (LDM~\cite{rombach2022high}) as our primary replay generator, leveraging its compressed latent space to achieve both computational efficiency and high-fidelity synthesis.

\paragraph{Autoencoder Configuration:}

We employ a VQ-regularized autoencoder~\cite{rombach2022high} with a downsampling factor of $f=4$ (referred to as \texttt{vq-f4}). The encoder compresses each $256 \times 256 \times 3$ input image into a latent feature map of size $64 \times 64 \times 3$. The autoencoder uses 128 base channels with channel multipliers of $[1, 2, 4]$ and includes two residual blocks at each resolution. The codebook contains $n_{embed}=8192$ entries, each with an embedding dimension of 3.

\paragraph{Diffusion and Network Configuration:}

The diffusion process is modeled in the latent space using a UNet~\cite{ronneberger2015u} architecture with the following specifications:

\begin{itemize}
\item \textbf{Backbone:} A time-conditional UNet with 224 base channels is used as the denoising backbone, comprising two residual blocks per level and channel multipliers of $[1, 2, 3, 4]$.
\item \textbf{Attention:} Spatial attention is incorporated at resolutions corresponding to downsampling factors of $[8, 4, 2]$ (i.e., 8, 16, and 32), with each attention head configured with 32 channels.
\item \textbf{Noise Schedule:} A linear noise schedule is used in the forward diffusion process, ranging from $\beta_{start}=0.0015$ to $\beta_{end}=0.0195$, and the model is optimized over $T=1000$ timesteps with an $L_2$ reconstruction objective.
\item \textbf{Optimization:} Training is performed using a base learning rate of $2.0 \times 10^{-6}$ to ensure stable convergence.
\end{itemize}

\paragraph{Sampling Configuration.}

During the generative replay phase, we employ the Denoising Diffusion Implicit Model (DDIM~\cite{song2020denoising}) sampler to accelerate synthesis. The sampling process is configured with the following parameters:

\begin{itemize}
\item \textbf{Sampling Steps:} We perform inference with 250 steps. Although the model is trained over 1000 timesteps, a strided sampling schedule is used to substantially speed up generation while preserving high fidelity.

\item \textbf{Stochasticity ($\eta$):} We use the DDIM sampler with the stochasticity parameter set to $\eta = 1.0$. This configuration effectively implements strided DDPM~\cite{ho2020denoising} sampling, ensuring that the generative replay preserves diversity comparable to the original training distribution.
\end{itemize}

\section{Further Analysis and Ablations}

\subsection{Results on Protocol 2}
The main paper details our experiments on the more realistic Protocol 1. To ensure fair alignment with established baselines, we further evaluate our method on Protocol 2. The quantitative results for this benchmark-aligned setup are presented in Tab. \ref{tab:p2_comparison}.

\begin{table}[h!]
\centering
\caption{Quantitative comparison (AUC) on Protocol 2. L-Bound denotes vanilla incremental learning without any strategy. }
\footnotesize
         \setlength{\tabcolsep}{2pt} 
\begin{tabular}{l@{\hspace{3pt}}  c | c c c c c} 
\toprule
Method & Task & DDPM & FF++ & DFDCP & CDF & Avg. \\ 
\toprule

\multirow{4}{*}{L-Bound} 
& T1 & 99.99 & - & - & - & 99.99 \\
& T2 & 69.60 & 95.16 & - & - & 82.38 \\
& T3 & 60.58 & 71.28 & 92.38 & - & 74.75 \\
& T4 & 53.39 & 65.92 & 82.23 & 99.77 & 75.33 \\
\midrule

\multirow{4}{*}{LwF} 
& T1 & 99.99 & - & - & - & 99.99 \\
& T2 & 81.01 & 85.02 & - & - & 83.02 \\
& T3 & 76.12 & 65.07 & 93.91 & - & 78.37 \\
& T4 & 61.71 & 61.14 & 81.35 & 98.18 & 75.60 \\
\midrule

\multirow{4}{*}{iCaRL} 
& T1 & 99.99& - & - & - & 99.99 \\
& T2 & 96.56 & 91.02 & - & - & 93.79 \\
& T3 & 91.71 & 80.07 & 90.95 & - & 87.58 \\
& T4 & 91.79 & 73.93 & 87.10 & 99.09 & 87.98 \\
\midrule

\multirow{4}{*}{DFIL} 
& T1 & 99.99 & - & - & - & 99.99 \\
& T2 & 97.17 & 93.31 & - & - & 95.24 \\
& T3 & 93.51 & 71.27 & 91.64 & - & 85.47 \\
& T4 & 92.33 & 68.10 & 80.74 & 99.63 & 85.20 \\
\midrule

\multirow{4}{*}{SUR-LID} 
& T1 & 99.99 & - & - & - & 99.99 \\
& T2 & 99.35 & 92.36 & - & - & 95.85 \\
& T3 & 99.13 & 81.51 & 92.13 & - & 90.92 \\
& T4 & 98.16 & 78.16 & 86.37 & 99.18 & 90.47 \\
\midrule

\multirow{4}{*}{\textbf{Dual-CARE (Ours)}} 
& T1 & 99.99 & - & - & - & 99.99 \\
& T2 & 99.88 & 94.51 & - & - & \textbf{97.20} \\
& T3 & 99.60 & 89.28 & 91.97 & - & \textbf{93.62} \\
& T4 & 98.87 & 85.90 & 92.04 & 98.12 & \textbf{93.73} \\

\bottomrule
\end{tabular}
\label{tab:p2_comparison}
\end{table}

\subsection{Comprehensive Ablation on Core Components}

In the main paper, we evaluate the ``Dual'' components separately to investigate their isolated performance. Here, we aim to investigate their mutual influence and the performance as a whole.
To systematically evaluate the contribution and synergy of our mechanisms, we extend the leave-one-out analysis from the main text to a comprehensive permutation of all three core components: Domain-aware Fast Training (DFT), Direct Gen-Real Supervision (DGS), and Relative Separation Loss (RSL). As shown in Tab. \ref{tab:detailed_ablation}, the results reveal a clear hierarchical dependency. Explicit label guidance from Gen-Real Sup. serves as the foundational anchor; omitting it leads to a severe performance collapse, as the detector fundamentally fails to retain historical real concepts. Building upon this established baseline boundary, the introduction of $\mathcal{L}_{rs}$ consistently provides a substantial performance lift by adaptively resolving the domain confusion between generated real and fake samples. Furthermore, the integration of DC-guided Gen ensures that the generative replay maintains stable and high-fidelity distributions across incremental tasks. Ultimately, the synergistic combination of all three components achieves the optimal balance between prior knowledge preservation and generative artifact mitigation.


\begin{table*}[htbp]
    \centering
    \caption{Comprehensive ablation study on the combinations of core components. The synergistic integration of all three modules yields the optimal incremental detection performance.}
    \label{tab:detailed_ablation}
    \begin{tabular}{ccc|cccccc}
    \toprule
        \textbf{DFT} & \textbf{DGS} & \textbf{RSL} & LDM & DFDCP & SDv21 & DDPM & DiT & Avg. \\
    \midrule
        \xmark & \xmark & \xmark & 89.47 & 81.44 & 95.91 & 53.15 & 61.53 & 76.30 \\
        \xmark & \xmark & \cmark & 94.11 & 80.63 & 95.40 & 53.21 & 58.62 & 76.39 \\
        \xmark & \cmark & \xmark & 92.06 & 85.28 & 96.64 & 94.65 & 87.16 & 91.15 \\
        \xmark & \cmark & \cmark & 95.42 & 83.61 & 94.16 & 96.38 & 94.99 & 92.91 \\
        \cmark & \xmark & \xmark & 94.94 & 78.98 & 96.42 & 50.12 & 58.98 & 75.88 \\
        \cmark & \xmark & \cmark & 92.99 & 78.60 & 97.51 & 56.29 & 55.42 & 76.16 \\
        \cmark & \cmark & \xmark & \textbf{99.39} & \textbf{88.40} & 96.10 & \textbf{98.38} & 86.05 & 93.66 \\
        \cmark & \cmark & \cmark & 98.45 & 87.36 & \textbf{97.71} & 95.28 & \textbf{98.07} & \textbf{95.37} \\
    \bottomrule
    \end{tabular}
\end{table*}

\subsection{Impact of Sample-wise Constraint in $\mathcal{L}_{rs}$}

\begin{figure}[t]
    \centering
    \includegraphics[width=1\linewidth]{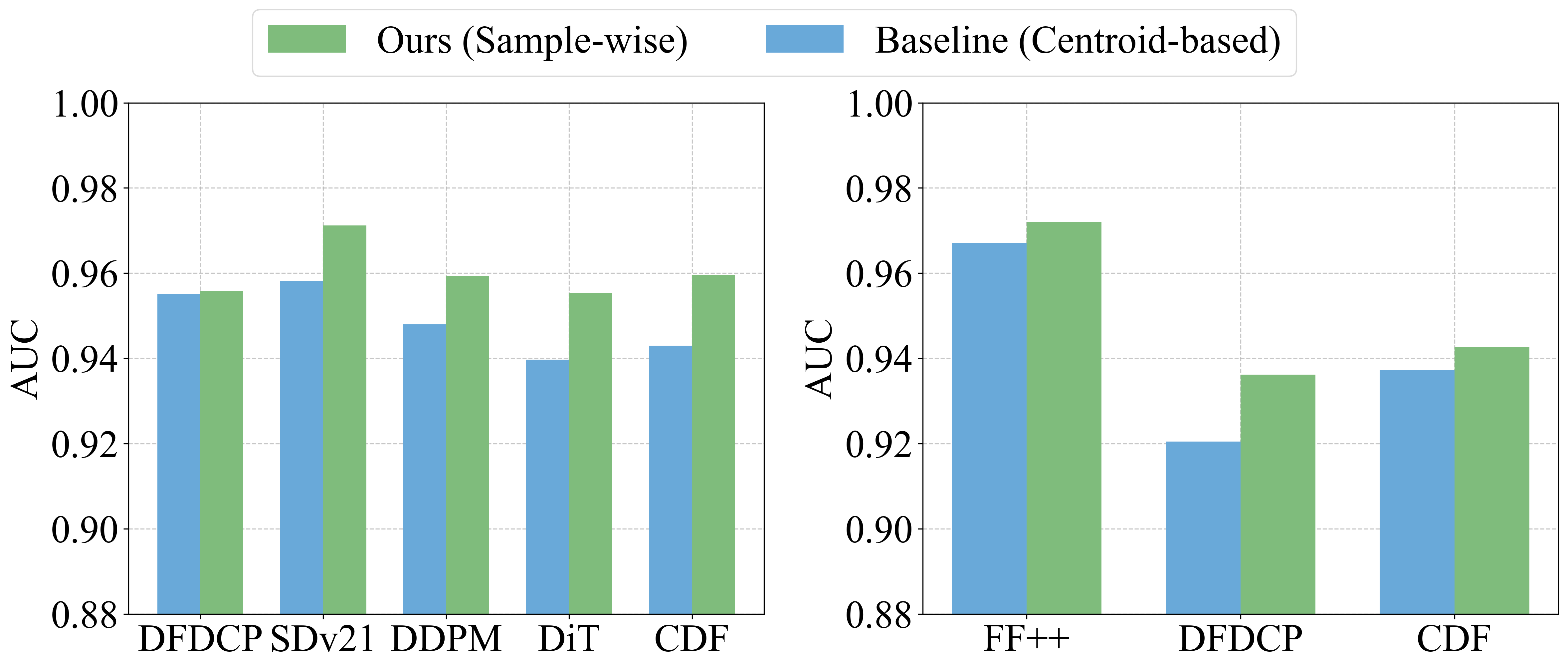}
    \caption{Ablation on Relative Separation Loss. The sample-wise formulation consistently outperforms the centroid-based variant across both Protocol 1 (Left) and Protocol 2 (Right) by preserving fine-grained structural diversity.}
    \label{fig:sample}
\end{figure}

To assess the importance of the fine-grained sample-wise constraint in our Relative Separation Loss ($\mathcal{L}_{rs}$), we compare it against a coarse-grained centroid-based variant. As shown in Fig. \ref{fig:sample}, the sample-wise formulation consistently outperforms the centroid-based strategy across all stages in both protocols. By enforcing separation at the instance level rather than relying on collapsed domain centers, our approach prevents local decision boundary distortions and preserves structural diversity. This fine-grained supervision proves crucial for mitigating catastrophic forgetting, particularly as domain confusion intensifies in later incremental stages.


\subsection{Robustness of Domain Confusion Score Normalization}

The Domain Confusion Score ($s_t$) is essential for dynamically balancing the supervision signals. To assess the method’s sensitivity to different normalization functions, we compare our default hyperbolic tangent (\texttt{tanh}) with Sigmoid and Linear Scaling (\texttt{d/5}). As shown in Table~\ref{tab:alpha_p2}, all variants produce reasonable results, but \texttt{tanh} yields the highest average AUC. This advantage stems from \texttt{tanh}’s smooth and bounded mapping of feature distances, which more effectively accommodates the varying magnitudes of domain shifts encountered throughout incremental learning.

\begin{table}[h]
\centering
\footnotesize
\caption{Performance comparison of normalization functions under Protocol 2.}
\setlength{\tabcolsep}{3.0pt} 
\begin{tabular}{l@{\hspace{3pt}} c | c c c c c c}
\toprule
Method & Task & DDPM & FF++ & DFDCP & CDF & Avg. & $s_t$ \\
\toprule

\multirow{4}{*}{tanh} 
& T1 & 99.99 & - & - & - & 99.99 &\\
& T2 & 99.88 & 94.51 & - & - & 97.20 & 0.9997\\
& T3 & 99.60 & 89.28 & 91.97 & - & 93.62 & 0.9999\\
& T4 & 98.87 & 85.90 & 92.04 & 98.12 & 93.73 & 0.9769\\
\midrule
\multirow{4}{*}{sigmoid}
& T1 & 99.99 & - & - & - & 99.99 & - \\
& T2 & 97.99 & 94.97 & - & - & 96.48 & 0.9875 \\
& T3 & 98.41 & 88.29 & 89.79 & - & 92.16 & 0.9971 \\
& T4 & 98.83 & 79.24 & 89.23 & 98.88 & 91.54 & 0.9013 \\
\midrule

\multirow{4}{*}{d/5}
& T1 & 99.99 & - & - & - & 99.99 & - \\
& T2 & 98.55 & 94.97 & - & - & 96.76 & 0.8077 \\
& T3 & 99.86 & 86.25 & 91.13 & - & 92.41 & 0.9999 \\
& T4 & 99.91 & 77.91 & 90.41 & 98.47 & 91.67 & 0.4694 \\

\bottomrule
\end{tabular}
\label{tab:alpha_p2}  
\end{table}

\subsection{Performance Comparison with Xception and ResNet34 Backbones}

\begin{figure}[htb]
    \centering
    \includegraphics[width=1\linewidth]{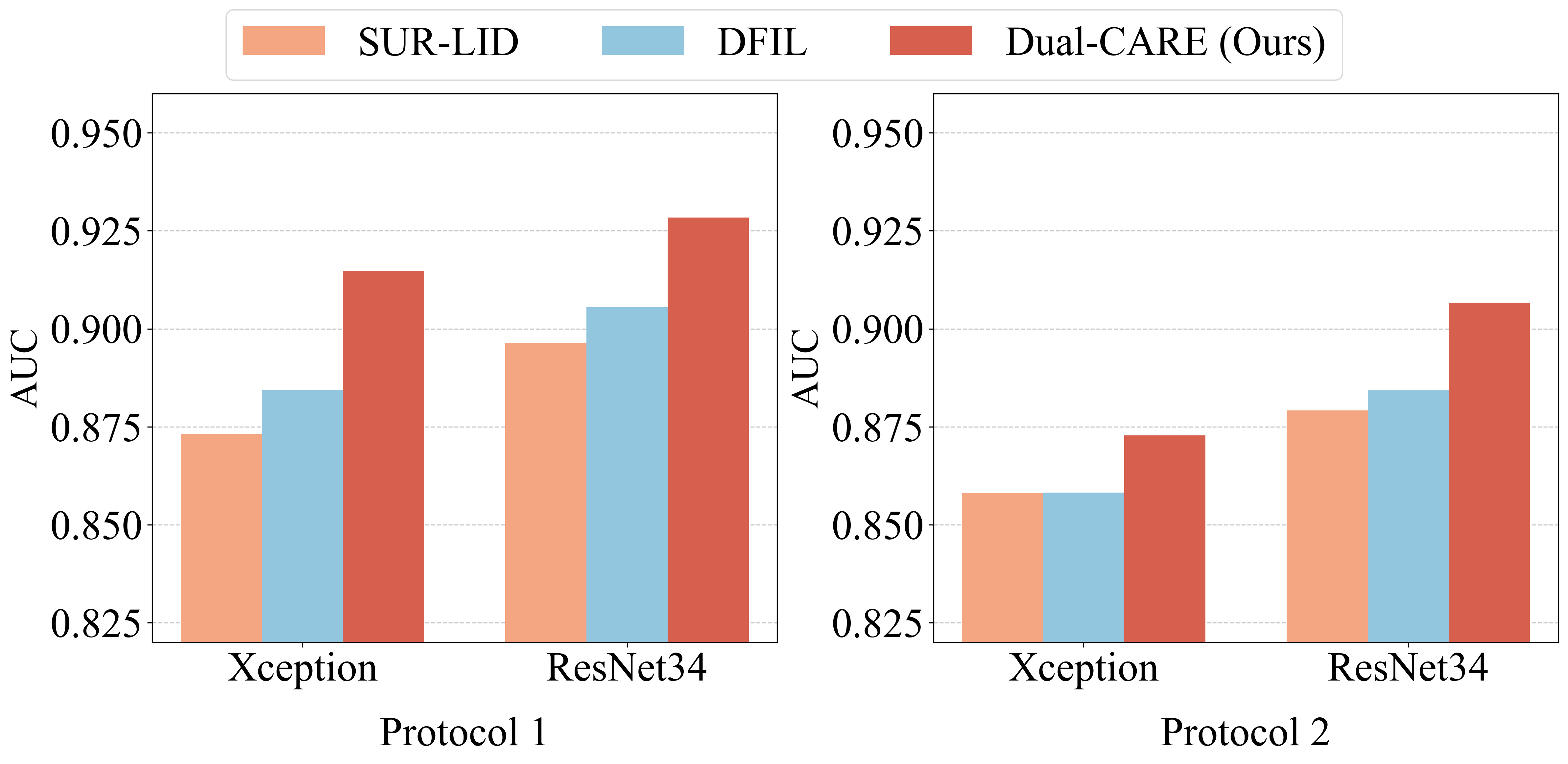} 
    \caption{Performance comparison with Xception and ResNet34 backbones.}
    
    \label{fig:backbone}
\end{figure}

To evaluate the generality of our framework beyond a specific feature extractor, we further test two additional backbones: Xception~\cite{xception} and ResNet34~\cite{resnet}. As shown in Fig.~\ref{fig:backbone}, we compare Dual-CARE with state-of-the-art incremental forgery detection methods, SUR-LID~\cite{cheng2025stacking} and DFIL~\cite{pan2023dfil}. Across both Protocol 1 and Protocol 2, Dual-CARE consistently outperforms all competitors, regardless of the backbone architecture. These results demonstrate that our Dual Confusion-Aware Regularization strategy is model-agnostic and remains robust across diverse network designs.

\subsection{Visualization of Domain-Safe Scenario}

\begin{figure}[h]
    \centering
    \includegraphics[width=1\linewidth]{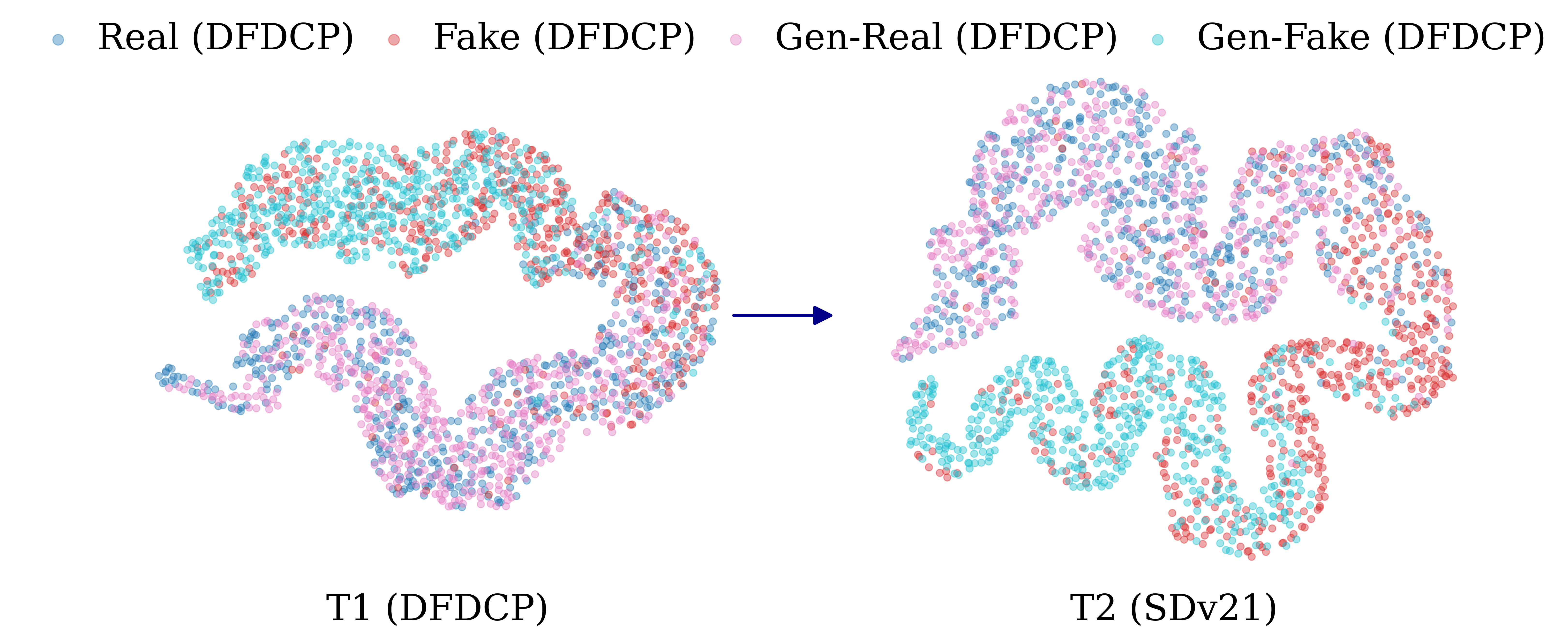} 
    \caption{UMAP~\cite{umap} visualization of the Domain-Safe scenario (DFDCP).}
    
    \label{fig:dfdcp}
\end{figure}

Complementing the domain-risky visualization in the main paper, we further illustrate a “domain-safe” scenario using the DFDCP dataset. As shown in Fig.~\ref{fig:dfdcp}, unlike the LDM case, the generated real samples for DFDCP initially align closely with the actual real samples, indicating minimal risk of domain confusion. Notably, this favorable alignment is well preserved even after learning the subsequent task (SDv21~\cite{sdv15}). This result demonstrates that our adaptive Dual-CARE strategy effectively identifies domain-safe samples and applies appropriate direct supervision to maintain their distributional integrity, rather than enforcing unnecessary separation.

\section{Sample Visualizations of Generative Replay}

\begin{figure}[h] 
    \centering
    \captionsetup{font=small} 
    
   (a) Stage 1: DiffusionFace (Real)\\
    \vspace{0.1cm}
    \begin{minipage}{0.19\linewidth}
        \includegraphics[width=\linewidth]{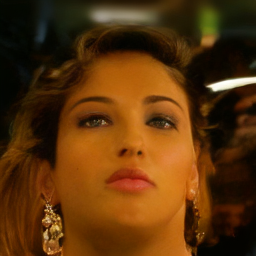} 
    \end{minipage}
    \hfill
    \begin{minipage}{0.19\linewidth}
        \includegraphics[width=\linewidth]{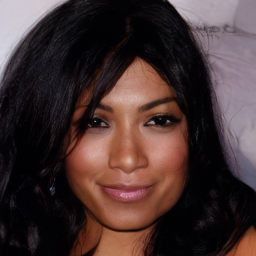}
    \end{minipage}
    \hfill
    \begin{minipage}{0.19\linewidth}
        \includegraphics[width=\linewidth]{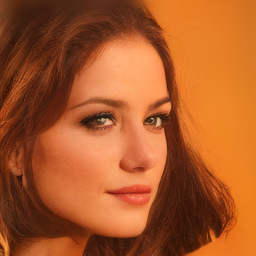}
    \end{minipage}
    \hfill
    \begin{minipage}{0.19\linewidth}
        \includegraphics[width=\linewidth]{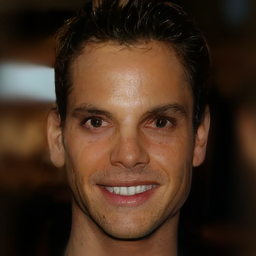}
    \end{minipage}
    
    \vspace{0.19cm} 
    
    (b) Stage 1: DiffusionFace (DDPM)\\
    \vspace{0.1cm}
    \begin{minipage}{0.19\linewidth}
        \includegraphics[width=\linewidth]{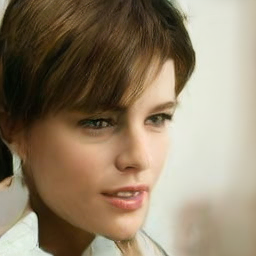}
    \end{minipage}
    \hfill
    \begin{minipage}{0.19\linewidth}
        \includegraphics[width=\linewidth]{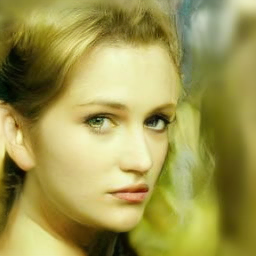}
    \end{minipage}
    \hfill
    \begin{minipage}{0.19\linewidth}
        \includegraphics[width=\linewidth]{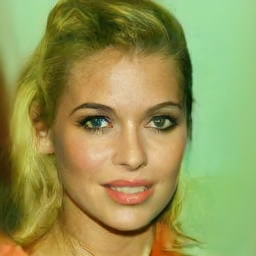}
    \end{minipage}
    \hfill
    \begin{minipage}{0.19\linewidth}
        \includegraphics[width=\linewidth]{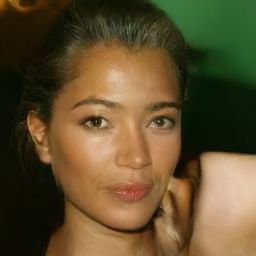}
    \end{minipage}
    
    \vspace{0.19cm}

    (c) Stage 2: Updated on FaceForensics++ (Real)\\
    \vspace{0.1cm}
    \begin{minipage}{0.19\linewidth}
        \includegraphics[width=\linewidth]{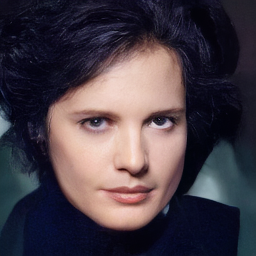}
    \end{minipage}
    \hfill
    \begin{minipage}{0.19\linewidth}
        \includegraphics[width=\linewidth]{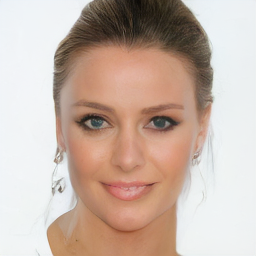}
    \end{minipage}
    \hfill
    \begin{minipage}{0.19\linewidth}
        \includegraphics[width=\linewidth]{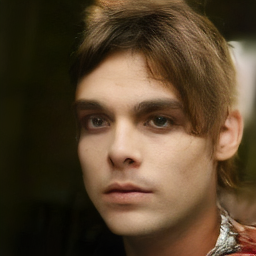}
    \end{minipage}
    \hfill
    \begin{minipage}{0.19\linewidth}
        \includegraphics[width=\linewidth]{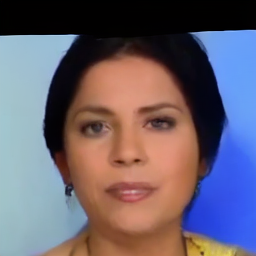}
    \end{minipage}
    
    \vspace{0.19cm}

    (d) Stage 2: Updated on FaceForensics++ (Fake)\\
    \vspace{0.1cm}
    \begin{minipage}{0.19\linewidth}
        \includegraphics[width=\linewidth]{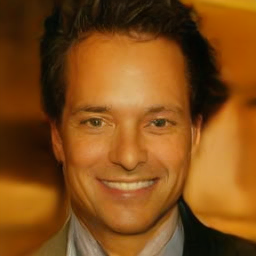}
    \end{minipage}
    \hfill
    \begin{minipage}{0.19\linewidth}
        \includegraphics[width=\linewidth]{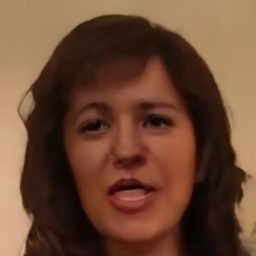}
    \end{minipage}
    \hfill
    \begin{minipage}{0.19\linewidth}
        \includegraphics[width=\linewidth]{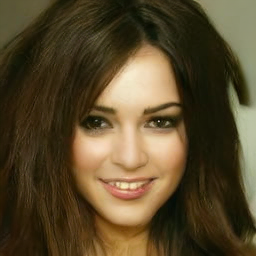}
    \end{minipage}
    \hfill
    \begin{minipage}{0.19\linewidth}
        \includegraphics[width=\linewidth]{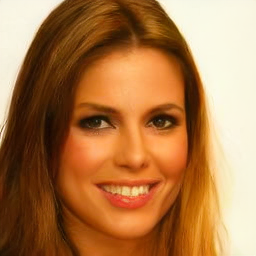}
    \end{minipage}
    
    \vspace{0.19cm}

    (e) Stage 3: Updated on DFDCP (Real)\\
    \vspace{0.1cm}
    \begin{minipage}{0.19\linewidth}
        \includegraphics[width=\linewidth]{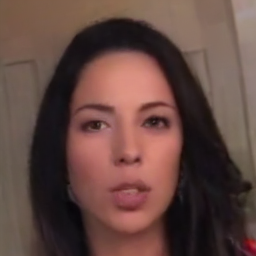}
    \end{minipage}
    \hfill
    \begin{minipage}{0.19\linewidth}
        \includegraphics[width=\linewidth]{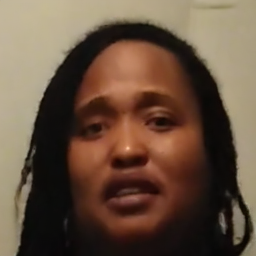}
    \end{minipage}
    \hfill
    \begin{minipage}{0.19\linewidth}
        \includegraphics[width=\linewidth]{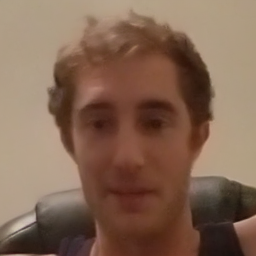}
    \end{minipage}
    \hfill
    \begin{minipage}{0.19\linewidth}
        \includegraphics[width=\linewidth]{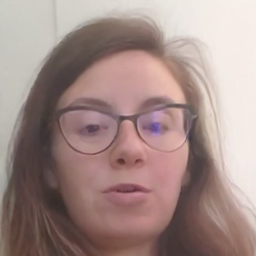}
    \end{minipage}
    
    \vspace{0.19cm}

    (f) Stage 3: Updated on DFDCP (Fake)\\
    \vspace{0.1cm}
    \begin{minipage}{0.19\linewidth}
        \includegraphics[width=\linewidth]{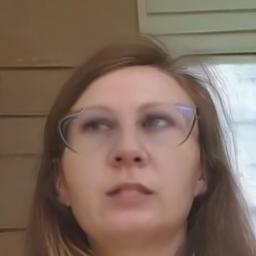}
    \end{minipage}
    \hfill
    \begin{minipage}{0.19\linewidth}
        \includegraphics[width=\linewidth]{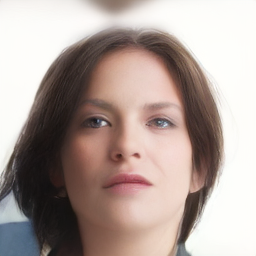}
    \end{minipage}
    \hfill
    \begin{minipage}{0.19\linewidth}
        \includegraphics[width=\linewidth]{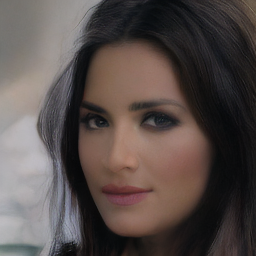}
    \end{minipage}
    \hfill
    \begin{minipage}{0.19\linewidth}
        \includegraphics[width=\linewidth]{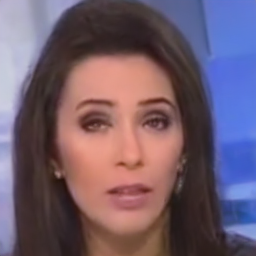}
    \end{minipage}

    \caption{LDM replay samples on (a-b) DiffusionFace, (c-d) FaceForensics++, and (e-f) DFDCP. A single updated generator retains high fidelity across tasks.}
    \label{fig:generated_replay_samples}
\end{figure}

We visualize replay samples across learning stages to demonstrate the robust memory retention of our mechanism. Our approach maintains a single, continuously updated model. As shown in Fig. \ref{fig:generated_replay_samples}, we track the generator's outputs from initial training on DiffusionFace (DDPM) through subsequent updates on FaceForensics++ (FF++) and DFDCP.

\section{Additional Experimental Results}
\label{sec:appendix_additional_results}



\subsection{Generative Replay Size and Diversity}

For fair comparison, Dual-CARE uses the same replay budget as
sample-replay baselines in all main experiments, i.e., 500 replay
samples per previous task. As shown in Fig.~\ref{fig:replay_analysis},
increasing the replay size consistently improves the incremental
detection performance, indicating that broader coverage of previous
task distributions helps alleviate catastrophic forgetting. Unlike
sample replay, which requires storing more historical images as the
replay size increases, generative replay can synthesize additional
samples from a fixed-size generator without increasing raw-data
storage. We further compare the diversity of sample replay and
generative replay under the same replay budget in
Tab.~\ref{tab:replay_diversity}. Generative replay achieves
consistently higher TCE, CLIP distance, and LPIPS scores than sample
replay, demonstrating its advantage in capturing richer
distributional, semantic, and perceptual variations. These results
further confirm that generative replay provides broader coverage of
historical task distributions than repeatedly using a fixed set of
stored samples.




\begin{table}[t]
\centering
\caption{Quantitative comparison of replay diversity under the
replay budget of 500 samples. TCE, CLIP Dist., and LPIPS measure
distributional, semantic, and perceptual diversity, respectively.
Higher values indicate greater diversity.}
\label{tab:replay_diversity}
\small
\setlength{\tabcolsep}{7pt}
\renewcommand{\arraystretch}{1.05}
\begin{tabular}{lccc}
\toprule
Replay Strategy
& TCE $\uparrow$
& CLIP Dist. $\uparrow$
& LPIPS $\uparrow$ \\
\midrule
Sample Replay
& 38.46
& 0.3786
& 0.5161 \\
Generative Replay
& \textbf{39.64}
& \textbf{0.4205}
& \textbf{0.5600} \\
\bottomrule
\end{tabular}
\end{table}

\subsection{DC Score and Distance Metric}

DC Score is quantified in the current-detector $f_t$ feature space because confusions between old and new tasks in IFFD are manifested in the representation geometry of $f_t$. Thus, our goal is not to estimate absolute distribution discrepancy, but to measure replay confusion perceived by $f_t$.
The dynamic feature space is appropriate for IFFD task transitions, as replay confusion also changes across tasks. Compared with a fixed distance metric, centroid distance under $f_t$ can more directly reflect the actual conflict from task $t$ to $t{+}1$.
More complex distribution metrics could be explored. However, our goal is to design an intuitive, lightweight, and detector-aligned control signal. DC Score only relies on feature centroid distance, introduces trivial overhead, and already consistently and robustly improves over fixed-$s_t$ variants.

As for L2 over Cosine Similarity: CosSim mainly captures angular consistency; L2 is additionally sensitive to feature magnitude and centroid displacement, making it more suitable for measuring confusion across evolving tasks.

\subsection{Fidelity of Non-Diffusion Replays}

Our goal is not to reproduce each historical manipulation process at the \textbf{pixel level}, but to \textbf{distill the common forgery-aware representations}, thus mitigating forgetting. In other words, replay samples do not need to exactly match the original samples, as long as they retain the discriminative forgery-aware information of previous tasks.

\textbf{Empirically}, although replays are generated by a unified LDM generator, we still achieve stable retention on traditional forgery datasets (e.g., DFDCP, CDF in Tab.~\ref{tab:main_comparison}). It suggests that the replay does not simply degenerate into ``LDM-style artifacts'', but can effectively preserve the common forgery information from prior tasks.

\subsection{Further Privacy Evidence}

We provide additional quantitative evidence for the privacy
preservation of generative replay in
Tab.~\ref{tab:supp-privacy}. First, we evaluate whether the
identities of generated replay samples can be matched to identities
in the original training set using ArcFace and CosFace. The low
identity-recall rates obtained by both recognition models indicate
that the replay generator does not directly reproduce training
identities. We further train a binary classifier to distinguish
generated replay samples from historical samples. Although
the classifier achieves high accuracy on the training split, its
performance drops to near-random accuracy on the held-out test
split, suggesting that generated and actual replay samples do not
exhibit distinguishable patterns. 
In summary, these results
show that generative replay preserves task-relevant distributions
without directly exposing or memorizing historical identities.

\begin{table}[t]
\centering
\caption{\small Quantitative privacy evaluation of generative replay.
ID Recall measures whether generated identities can be matched to
the training identities using face-recognition models.
Replay Distinguishment reports the accuracy of a classifier
for distinguishing generated replay from actual replay. Lower ID
Recall and test accuracy closer to chance indicate stronger privacy
preservation.}
\label{tab:supp-privacy}
\small
\setlength{\tabcolsep}{6pt}
\renewcommand{\arraystretch}{1.05}
\begin{tabular}{lc}
\toprule
Metric & Result (\%) \\
\midrule
ID Recall with ArcFace $\downarrow$ & 0.08 \\
ID Recall with CosFace $\downarrow$ & 0.05 \\
Replay Distinguishment -- Train Acc. & 96.88 \\
Replay Distinguishment -- Test Acc. $\downarrow$ & 46.50 \\
\bottomrule
\end{tabular}
\end{table}

\subsection{Protocol-Aligned Performance}

To facilitate direct comparison with existing IFFD methods, we
further evaluate Dual-CARE under the incremental protocol adopted
by prior work. As shown in Tab.~\ref{tab:protocol-align},
our method achieves competitive average performance while obtaining
the best results on DFD and CDF2, demonstrating its effectiveness
under established benchmark settings.

\begin{table}[t]
\centering
\caption{\small Quantitative comparison in terms of AUC (\%) under
the incremental protocol adopted by prior IFFD methods. The best
result in each column is highlighted in bold.}
\label{tab:protocol-align}
\small
\setlength{\tabcolsep}{4pt}
\renewcommand{\arraystretch}{0.9}
\begin{tabular}{lccccc}
\toprule
Method & FF++ & DFDCP & DFD & CDF2 & Avg. \\
\midrule
DMP   & \textbf{91.61} & 84.86 & 91.81 & 91.67 & \textbf{89.99} \\
DevDF & 90.71 & \textbf{90.31} & 93.12 & 85.15 & 89.82 \\
Ours  & 83.89 & 82.35 & \textbf{94.81} & \textbf{98.33} & 89.84 \\
\bottomrule
\end{tabular}
\end{table}

\subsection{Adaptation to IFFD}

Our strategy uses the DC Score to uniquely address domain confusion caused by rapid generator iterations in forgery detection. By specifically regulating domain-risky samples, it ensures clear separation between real and fake distributions, a task-specific adaptation missing in generic frameworks.

\subsection{Robustness}

\begin{figure}[H]
\centering
\includegraphics[width=0.95\linewidth]{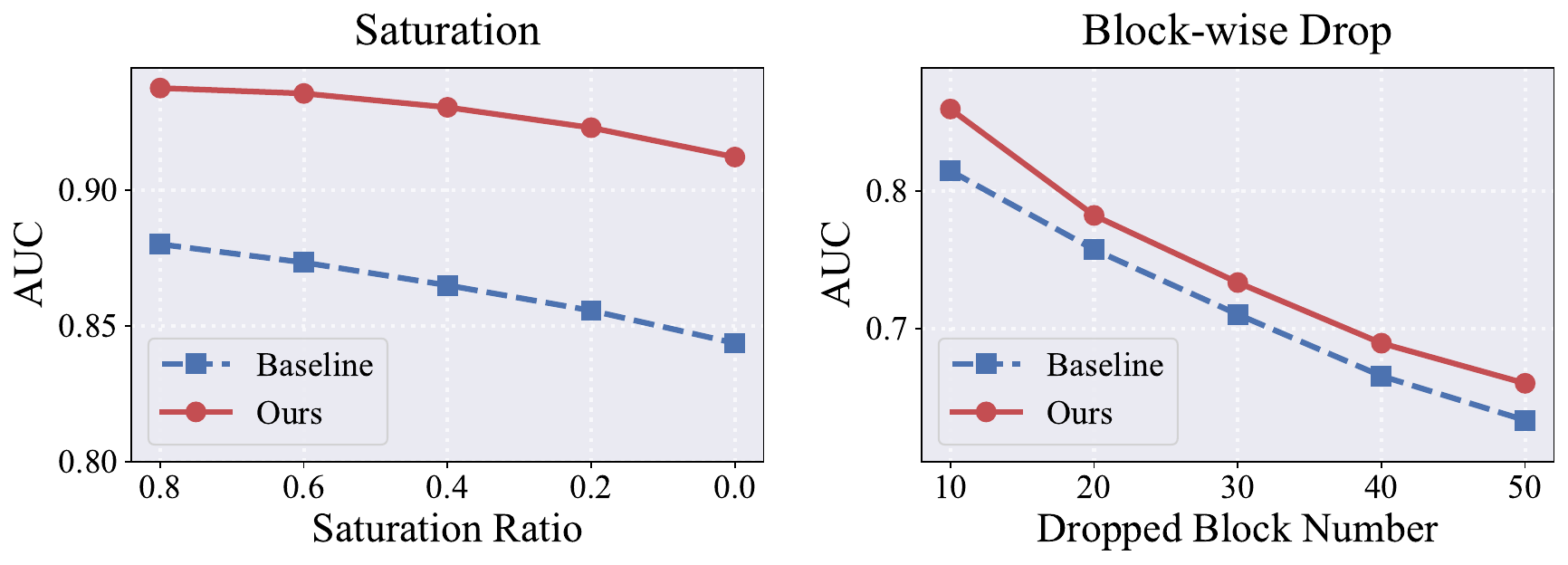}
\caption{\small Robustness evaluation under saturation changes and
block-wise drop with different perturbation strengths.}
\label{fig:robustness}
\end{figure}

We evaluate the robustness of Dual-CARE under common image
perturbations, including saturation changes and block-wise drop.
As shown in Fig.~\ref{fig:robustness}, our method maintains stable
detection performance across different perturbation strengths,
demonstrating its resistance to practical image degradations and
local information loss.

\subsection{Cross-domain Performance}

We further evaluate cross-domain generalization on several unseen
forgery datasets. As reported in Tab.~\ref{tab:generalization},
Dual-CARE consistently outperforms SUR-LID across all datasets,
demonstrating that the proposed confusion-aware replay strategy not
only preserves prior knowledge but also improves generalization to
unseen forgery distributions.

\begin{table}[H]
\centering
\caption{\small Cross-domain generalization performance in terms of
AUC (\%) on unseen forgery datasets. The best results are highlighted
in bold.}
\label{tab:generalization}
\small
\setlength{\tabcolsep}{4pt}
\renewcommand{\arraystretch}{0.9}
\begin{tabular}{lcccc}
\toprule
Method & DFD & DDIM & UADFV & WDF \\
\midrule
SUR-LID & 73.31 & 91.20 & 95.76 & 68.88 \\
Ours    & \textbf{77.97} & \textbf{92.45} & \textbf{96.25} & \textbf{70.73} \\
\bottomrule
\end{tabular}
\end{table}

\subsection{RS Loss over Sample-level Loss}

\begin{figure}[H]
\centering
\includegraphics[width=0.9\linewidth]{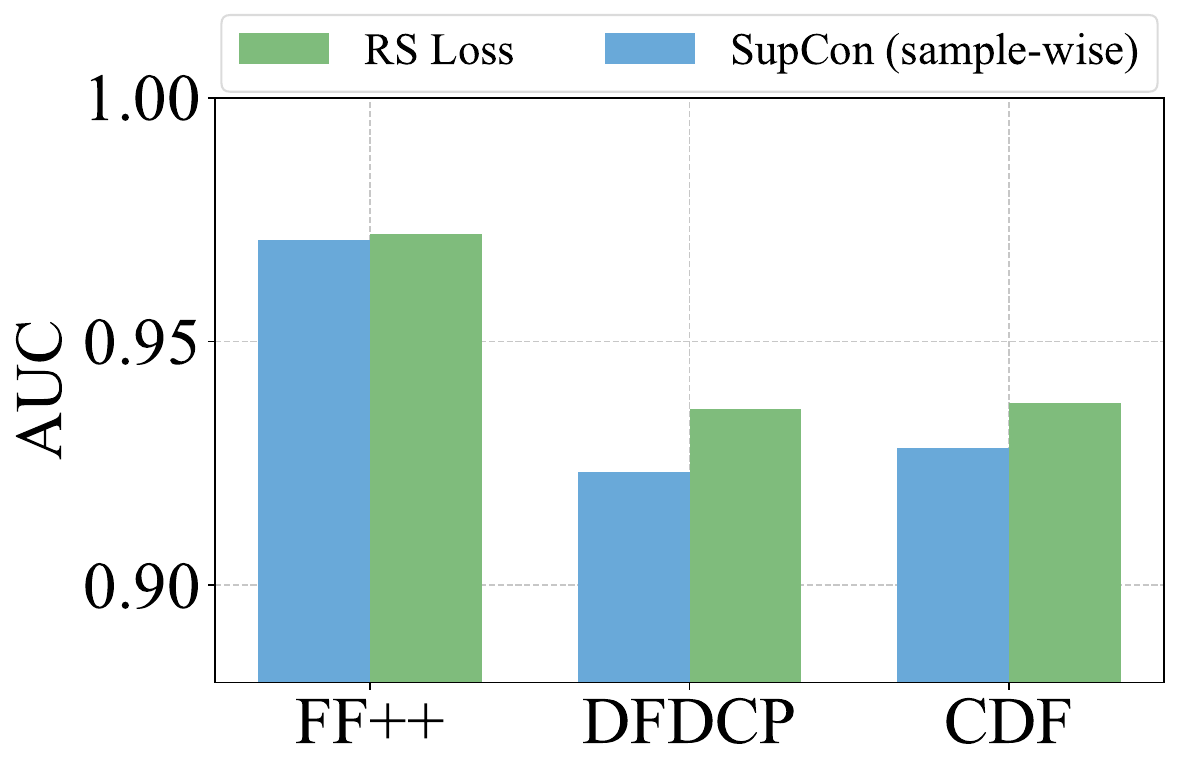}
\caption{\small Comparison between RS Loss and SupCon. RS Loss exhibits better domain-level feature aggregation and separation under incremental learning.}
\label{fig:supcon}
\end{figure}

RS Loss is distribution-level to measure overall confusion. It is less affected by sample-wise outliers and noisy replay artifacts, and is therefore better suited to capture the overall drift among domains. In Fig.~\ref{fig:supcon}, comparisons between RS Loss and SupCon demonstrate our claims.

\end{document}